\documentclass[Afour,sageh,times]{sagej}

\usepackage{moreverb,url}

\usepackage[colorlinks,bookmarksopen,bookmarksnumbered,citecolor=red,urlcolor=red]{hyperref}

\newcommand\BibTeX{{\rmfamily B\kern-.05em \textsc{i\kern-.025em b}\kern-.08em
T\kern-.1667em\lower.7ex\hbox{E}\kern-.125emX}}

\def\volumeyear{2019}
\usepackage{makeidx}         %
\usepackage{graphicx}        %
\usepackage{float}
\usepackage{enumerate}
\usepackage{subfig}
\usepackage{sidecap}
\usepackage{amssymb}
\usepackage{amsmath}
\usepackage{bbm}
\usepackage{mathtools}
\newcommand{\previouslyrevised}[1]{{\leavevmode\color{black}#1}}

\DeclareMathOperator*{\minimize}{minimize}

\newcommand{\xrob}{x_{_\mathrm{R}}}
\newcommand{\urob}{u_{_\mathrm{R}}}
\newcommand{\xr}{p_{x_\mathrm{R}}}
\newcommand{\yr}{p_{y_\mathrm{R}}}
\newcommand{\psir}{\psi_{_\mathrm{R}}}
\newcommand{\Uxr}{U_{x_\mathrm{R}}}
\newcommand{\Uyr}{U_{y_\mathrm{R}}}
\newcommand{\rr}{r_{_\mathrm{R}}}

\newcommand{\dxrob}{\dot{x}_{_\mathrm{R}}}

\newcommand{\Fx}{F_x}
\newcommand{\Fxf}{F_{x_\mathrm{f}}}
\newcommand{\Fxr}{F_{x_\mathrm{r}}}
\newcommand{\Fyf}{F_{y_\mathrm{f}}}
\newcommand{\Fyr}{F_{y_\mathrm{r}}}
\newcommand{\Fzf}{F_{z_\mathrm{f}}}
\newcommand{\Fzr}{F_{z_\mathrm{r}}}

\newcommand{\xhum}{x_{_\mathrm{H}}}
\newcommand{\uhum}{u_{_\mathrm{H}}}
\newcommand{\xh}{p_{x_\mathrm{H}}}
\newcommand{\yh}{p_{y_\mathrm{H}}}
\newcommand{\psih}{\psi_{_\mathrm{H}}}
\newcommand{\vh}{v_{_\mathrm{H}}}

\newcommand{\Xrel}{x_{_\mathrm{\rel}}}
\newcommand{\rel}{\mathrm{rel}}
\newcommand{\xrel}{p_{x_\rel}}

\newcommand{\yrel}{p_{y_\rel}}

\newcommand{\psirel}{\psi_{_\rel}}

\newcommand{\frel}{f_{_\rel}}

\newcommand{\HJI}{{_\mathrm{HJI}}}
\newcommand{\WALL}{{_\mathrm{WALL}}}

\begin{document}

\runninghead{Karen Leung, Edward Schmerling, et al.}

\title{On Infusing Reachability-Based Safety Assurance within Planning Frameworks for Human-Robot Vehicle Interactions}

\author{
Karen Leung\affilnum{1},
Edward Schmerling\affilnum{1},
Mengxuan Zhang\affilnum{1},
Mo Chen\affilnum{2},
John Talbot\affilnum{1},
J. Christian Gerdes\affilnum{1}, and
Marco Pavone\affilnum{1}}

\affiliation{\affilnum{1}Stanford University, Stanford, CA 94305, USA\\
\affilnum{2}Simon Fraser University, Burnaby, BC V5A 1S6, Canada}
\corrauth{Marco Pavone, 496 Lomita Mall, Rm. 261 Stanford, CA 94305
\email{pavone@stanford.edu}}

\begin{abstract}
Action anticipation, intent prediction, and proactive behavior are all desirable characteristics for autonomous driving policies in interactive scenarios. Paramount, however, is ensuring safety on the road---a key challenge in doing so is accounting for uncertainty in human driver actions without unduly impacting planner performance.
This paper introduces a minimally-interventional safety controller operating within an autonomous vehicle control stack with the role of ensuring collision-free interaction with an externally controlled (e.g., human-driven) counterpart while respecting static obstacles such as a road boundary wall. We leverage reachability analysis to construct a real-time (100Hz) controller that serves the dual role of (i) tracking an input trajectory from a higher-level planning algorithm using model predictive control, and (ii) assuring safety by maintaining the availability of a collision-free escape maneuver as a persistent constraint regardless of whatever future actions the other car takes. A full-scale steer-by-wire platform is used to conduct traffic weaving experiments wherein two cars, initially side-by-side, must swap lanes in a limited amount of time and distance, emulating cars merging onto/off of a highway. We demonstrate that, with our control stack, the autonomous vehicle is able to avoid collision even when the other car defies the planner's expectations and takes dangerous actions, either carelessly or with the intent to collide, and otherwise deviates minimally from the planned trajectory to the extent required to maintain safety.
\end{abstract}


\maketitle

\section{Introduction}\label{sec:intro}

Decision-making and control for mobile robots is typically stratified into levels.
A high-level planner, informed by representative yet simplified dynamics of a robot and its environment, might be responsible for selecting an optimal, yet coarse trajectory plan, which is then implemented through a low-level controller that respects more accurate models of the robot's dynamics and control constraints.
While additional components may be required to flesh out a robot's full control stack from model to motor commands, selecting the right ``division of responsibilities'' is fundamental to system design.

One consideration that defies clear classification, however, is how to ensure a mobile robot's safety when operating in close proximity with a rapidly evolving and stochastic environment.
Safety is a function of uncertainty in both the robot's dynamics and those of its surroundings; high-level planners typically do not replan sufficiently rapidly to ensure split-second reactivity to threats, yet low-level controllers are typically too short-sighted to ensure safety beyond their local horizon.

Human-robot interactions are an unavoidable aspect of many modern robotic applications and ensuring safety for these interactions is critical, especially in applications such as autonomous driving where collisions may lead to life-threatening injury.
However, ensuring safety within the planning and control framework can be very challenging due to the uncertainty in how humans may behave.
To quantify this uncertainty, robots often rely on generative models of human behavior in order to inform their planning algorithms \citep{SadighSastryEtAl2016c,SchmerlingLeungEtAl2018}, thereby enabling more efficient and communicative interactions.
In general, under nominal operating conditions that reflect the modeling assumptions, these model-based probabilistic planners can offer high performance (e.g., minimizing time and control effort for the robot). However, these planners alone are typically insufficient for ensuring absolute safety because (i) they depend on \emph{probabilistic} models of human behavior and thus safety is not enforced deterministically, (ii) dangerous but low-likelihood events may not be adequately captured in the human behavior prediction model, and (iii) reasoning about these probabilistic behavior models is typically too computationally expensive for the planners to react in real time when humans strongly defy expectations and/or diverge from modeling assumptions.

In this work we implement a control stack for a full-scale autonomous car (the ``robot'') engaging in close proximity interactions with a human-controlled vehicle (the ``human''). Our control stack aims to stay true to planned trajectories from a high-level planner since freedom of motion is essential for the planner to carry out the driving task while conveying future intent to the other vehicle. At the same time, we allow the robot to deviate from the desired trajectory to the point that is necessary to maintain safety.
Our primary tool for designing a controller that does not needlessly impinge upon the planner's choices is \emph{Hamilton-Jacobi (HJ) backward reachability}.
We provide a brief overview of the reachability analysis literature relevant to our work in Section \ref{sec:BRS}.

\section{Related Work}

For high-level planners, safety is often incentivized, but not strictly enforced as a hard constraint. For example, safety is often part of the objective function when selecting optimal plans, or represented via artificial potential fields \citep{WolfBurdick2008,SadighSastryEtAl2016c,SchmerlingLeungEtAl2018}. Although these approaches are designed to account for interactive scenarios in which another sentient agent is a key environmental consideration, they contain competing objectives (i.e., collision avoidance vs. goal-oriented performance), often do not plan at a sufficiently high rate to account for rapidly-changing environments, and ultimately provide no theoretical framework for ensuring safety for both the human and the robot.

Another common approach to finding collision-free plans is to use \emph{forward} reachability. The idea is to fix a time horizon and compute the set of states where the other agents could possibly be in the future and plan trajectories for the robot that avoid this set \citep{AlthoffDolan2011,LiuRoehmEtAl2017,LorenzettiChenEtAl2018}. Although this gives a stronger sense of safety, this is only practical for short time horizons otherwise it will lead to overly conservative robot behaviors or even planning problem infeasibility as the set to avoid grows. In general, approaches that use forward reachability results in very conservative results, especially for interactive scenarios where freedom of motion (e.g., nudging) is necessary to convey intent. These methods, as well as methods that enforce safety as an objective, are often subjected to model simplification, such as using a linear dynamics model instead of a nonlinear one, or making simplifying assumptions for computational tractability. However, model simplifications can lead to overly conservative or imprecise results which may impede performance.

Safety can also be introduced at the low-level which often obeys higher fidelity dynamics models. A common approach in ensuring safe low-level controls is to use reactive collision avoidance techniques---the robot is normally allowed to apply any control, but switches to an avoidance controller when near safety violation.
Examples of this general approach include HJ backward reachability-based controllers which have been applied in such a switched fashion \citep{FisacAkametaluEtAl2018,ChenTomlin2018,BajcsyBansalEtAl2019} and have proven to be effective for avoiding other interacting agents and static obstacles.
HJ reachability has been studied extensively and applied successfully in a variety of safety-critical interactive settings \citep{MitchellBayenEtAl2005,BokanowskiForcadelEtAl2010,GattamiAlAlamEtAl2011,MargellosLygeros2011,ChenHuEtAl2017,DhinakaranChenEtAl2017} due to its flexibility with respect to system dynamics, and its optimal (i.e., non-overly-conservative) avoidance maneuvers stemming from its equivalence to an exhaustive search over joint system dynamics.
Other approaches include using a precomputed emergency maneuver library \citep{AroraChoudhuryEtAl2015}. A key drawback to switching controllers, however, is that the performance goals considered by the high-level planner are completely ignored when the reactive controller steps in. For some cases, this may be acceptable and necessary, but in general, it is desirable to ensure safety without unduly impacting planner performance where possible.
Instead of switching to a different controller,  \cite{GrayGaoEtAl2013}, \cite{FunkeBrownEtAl2017}, and \cite{BrownFunkeEtAl2017} adapt the low-level online optimal controller to incorporate constraints that avoid static obstacles. The approach taken in \cite{GrayGaoEtAl2013} aims to be minimally interventional, however, they only consider cases with disturbance uncertainty rather than environmental uncertainty stemming from the system interacting with other human agents.
These approaches are able to strike the optimal balance between tracking performance and safety (i.e., optimizing performance subject to safety as a hard constraint), but these approaches as presented are effective for avoiding static obstacles only.

Inspired by the non-conservative nature and optimality of HJ reachability, and the performance of online optimal control for trajectory tracking, this work aims to infuse reachability-based safety assurance into the low-level controller such that when the robot is near safety violation, the robot is able to simultaneously maintain safety and follow the high-level plans without unduly impacting performance.
To the best of our knowledge there has not been any work that explicitly addresses the integration of reachability-based safety controllers as a component within a robot's control stack, i.e., with safety as a constraint upon a primary planning objective.

\begin{figure*}[ht]
    \centering
    \subfloat[The human's dynamics are propagated \emph{forward} in time. The robot plans trajectories that stay outside of the human's FRS (shaded region).]{
    \label{fig:FRS illustration}
    \includegraphics[width=0.33\textwidth]{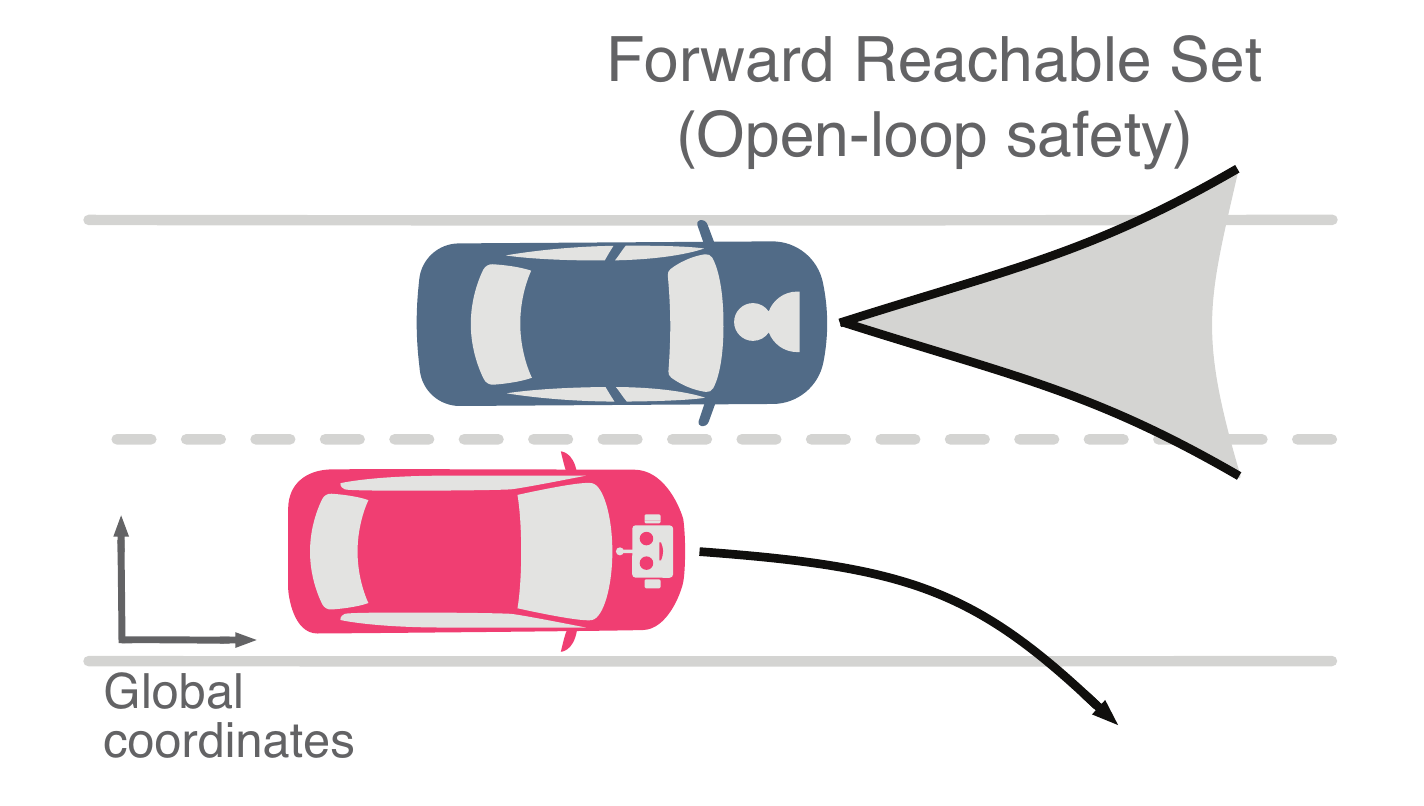}} $\qquad$ \subfloat[Joint dynamics are propagated \emph{backwards} in time. The BRS (shaded region) is not overly-conservative even for long time horizons because the robot can react if the human was to swerve (see right). The robot is in an unsafe state if the human enters the BRS, defined in the relative state space.]{
    \label{fig:BRS illustration}
    \includegraphics[width=0.58\textwidth]{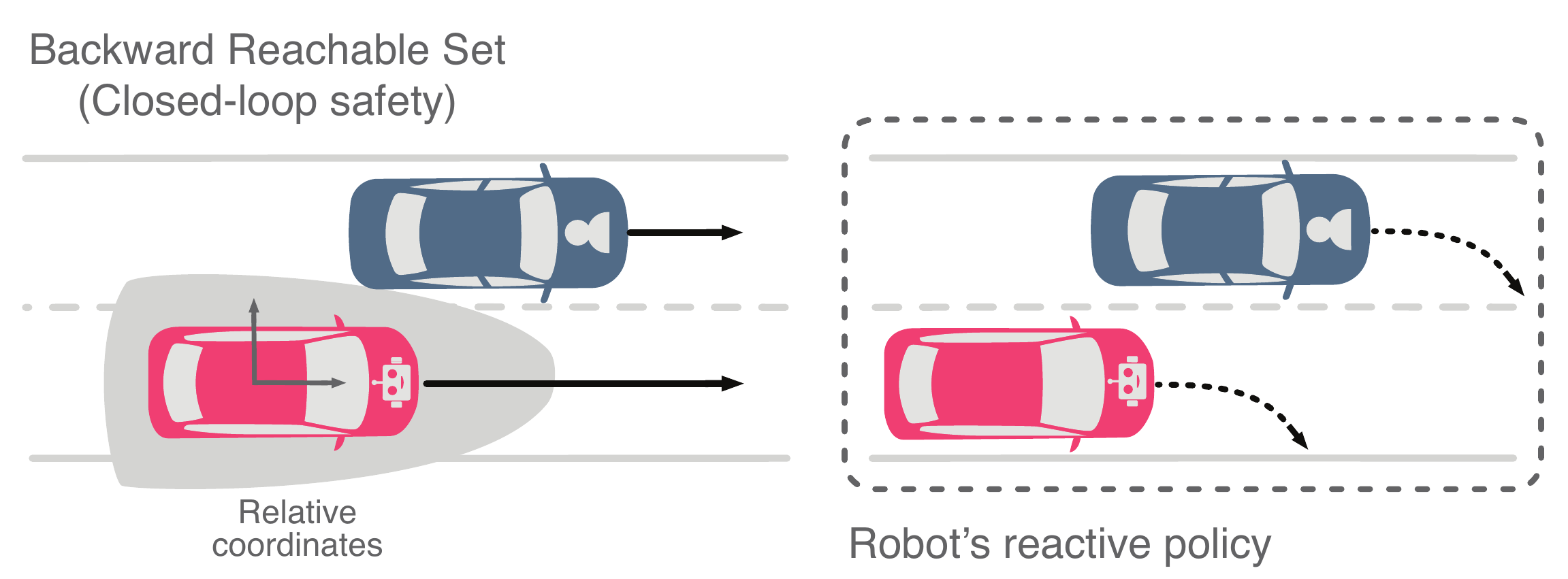} }
\caption{Illustration of forward and backward reachability, and how they are commonly used to ensure safe planning and controls. The robot and human are depicted in red and blue respectively.}
\label{fig:reachability}
\end{figure*}

\noindent \textit{\textbf{Statement of Contributions:}} The contributions of this paper are twofold. First, we propose a method for formally incorporating reachability-based safety within an existing optimization-based control framework. The main insight that enables our approach is the recognition that, near safety violation, the set of safety-preserving controls often contains more than just the optimal avoidance control. Instead of directly applying this optimal avoidance control when prompted by reachability considerations, as in a switching control approach, we quantify the set of safety-preserving controls and pass it to the broader control framework as a constraint. Our intent is to enable minimal intervention against the direction of a higher-level planner when evasive action is required. Second, we evaluate the benefits, performance, and trade-offs of this safe control methodology in the context of a probabilistic planning framework for the traffic weaving scenario studied at a high level in \cite{SchmerlingLeungEtAl2018}, wherein two cars, initially side-by-side, must swap lanes in a limited amount of time and distance. Experiments with a full-scale steer-by-wire vehicle reveal that our combined control stack achieves better safety than applying a tracking controller alone to the planner output, and smoother operation (with similar safety) compared with a switching control scheme; in our discussion we provide a roadmap towards improving the level of safety assurance in the face of practical considerations such as unmodeled dynamics, as well as towards generalizations of the basic traffic weaving scenario.

A preliminary version of this work appeared at the 2018 International Symposium on Experimental Robotics (ISER). In this revised and extended version, we provide the following additional contributions: (i) an extension to the control framework to account for static obstacles, (ii) more exposition on the vehicle dynamics models used and a deeper discussion on the underlying modeling assumptions, (iii) an empirical study of the safety-efficiency trade-off for our approach, and (iv) additional experimental results including trials with a static road boundary wall and a comparison with a baseline approach.

\begin{figure*}[ht]
    \centering
    \includegraphics[width=\textwidth]{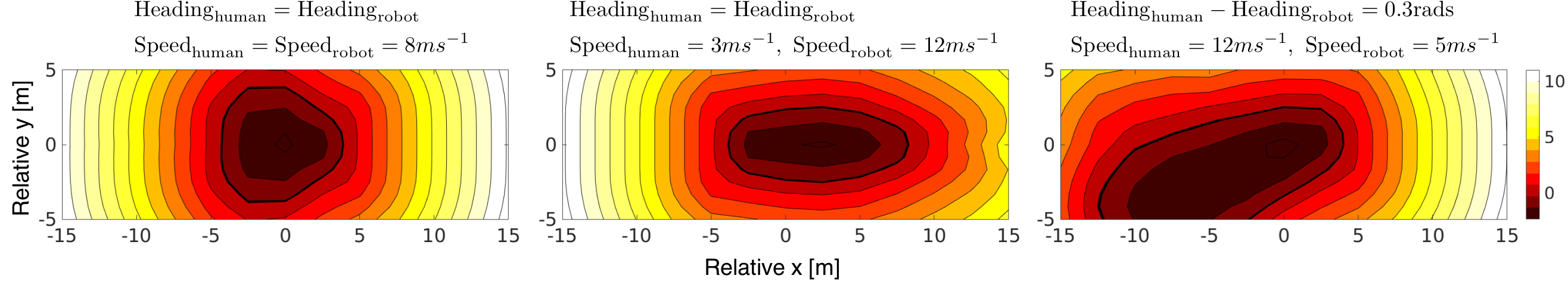}
    \caption{Contour plots of slices of the HJI value function $V$ computed from the relative dynamics (see Equation \eqref{eqn:relative dynamics}) and choice of signed distance as the terminal value function. Slices show $V$ as a function of relative $x$ and $y$ position (see Figure \ref{fig:relative frame} for an illustration of the human-robot relative pose) with all other states held fixed. The zero-level set, outlined with a thick black line, is the backwards reachable set from collision states (i.e., the avoid set).}
    \label{fig:collision_avoid_set}
\end{figure*}

\previouslyrevised{
\begin{figure}[tb]
    \centering
    \includegraphics[width=0.4\textwidth]{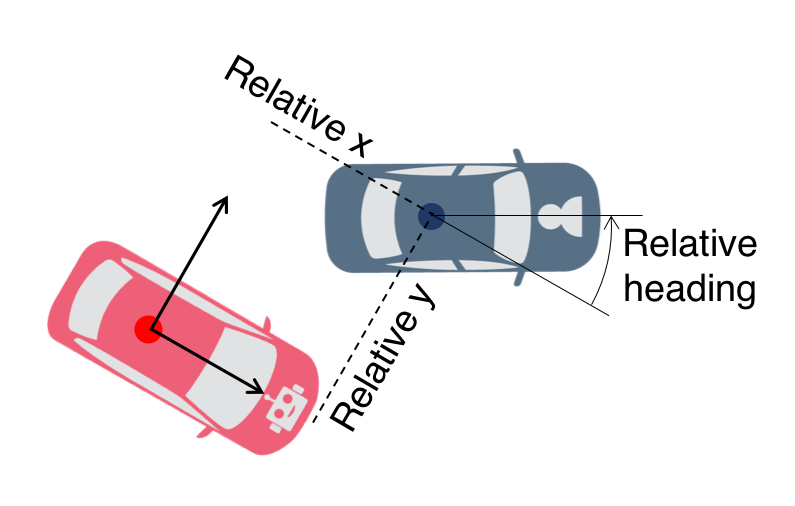}
    \caption{The relative human-robot coordinate frame and relative pose between the robot (red) car and human (blue) car. The frame is centered and rotated about the robot's pose.}
    \label{fig:relative frame}
\end{figure}
}

\section{Background: Hamilton-Jacobi Reachability}
\label{sec:BRS}
In this section, we provide a brief overview of reachability analysis as a tool for constructing safe trajectory plans, and introduce HJ backward reachability concepts relevant to this work.

\subsection{Overview: Reachability Analysis}
Given a dynamics model governing a robotic system incorporating control and disturbance inputs, reachability analysis is the study of the set of states that the system can reach from its initial conditions. It is often used for formal verification as it can give guarantees on whether or not the evolution of the system will be safe, i.e., whether the reachable set includes undesirable outcomes. Reachability analysis can be divided into two main paradigms: (i) forward reachability and (ii) backward reachability.

\noindent \emph{Forward Reachability Analysis: }The \emph{forward reachable set} (FRS) is the set of states that the system could potentially be in after some time horizon $t$. This is computed by propagating the dynamics combined with all feasible control sequences and disturbances forward in time. When considering the interaction between two agents, for example a human and a robot, the forward reachable set is computed for the human and the robot plans to avoid this set to ensure collision-free trajectories (see Figure~\ref{fig:FRS illustration}). This \emph{open-loop} mentality, however, leads to an overly-conservative robot outlook. That is, while considering actions in the present, the robot does not incorporate the possibility that its future observations of where the human goes might influence how much it actually needs to take avoidance actions; in short the robot's plan to avoid the FRS does not incorporate closed-loop feedback. To reduce the over-conservative nature of forward reachability, the time horizon $t$ for which the FRS is computed over is typically kept small and is recomputed frequently. This approach has been found to be effective in finding collision-free trajectories \citep{AlthoffDolan2014}, but it is difficult to extend to interactive scenarios where there are more uncertainties in the rapidly changing environment.
Aside from the overly-conservative nature of FRS, the key drawback with using forward reachability is that safety (i.e., avoiding the FRS) hinges on the planner's capabilities and operating frequency. Even if computing the FRS is instantaneous, the planner may still be unable to react to split-second threats.

\noindent \emph{Backward Reachability Analysis: }
Let the target set represent a set of undesirable states (e.g., collision states of the system).
The \emph{backward reachable set} (BRS) is the set of states that could result in the system being in the target set, assuming worst-case disturbances, after some time horizon $t$.
Specifically, the BRS represents the set of states
from which there does not exist a controller that can prevent the robot dynamics from being driven into the target set under worst-case disturbances within a time horizon $t$.
As such, to rule out such an eventuality, the BRS is treated as the ``avoid set''.
Critical differences between the FRS and BRS are that (i) the BRS is computed \emph{backwards} in time, and (ii) the BRS is computed assuming \emph{closed-loop} reactions to the disturbances.
Illustrated in Figure \ref{fig:BRS illustration}, for the case of relative dynamics between human- and robot-controlled vehicles, computation of the BRS takes into account the fact that the robot is able to react to the human at any time and in any state configuration (if the human was to swerve into the robot, the robot can swerve too to avoid a collision).
This leads to the BRS being less overly-conservative than the FRS.
In practice, the results of the BRS computation are cached via a look-up table, and at run-time, the optimal robot policy can be computed via a near-instant lookup of the reachability cache. As such, we can always compute optimal actions for the robot at any state configuration and \emph{regardless} of the high-level planner used.

We elect to use backward reachability because (i) its non-overly conservative nature stemming from the closed-loop computation ensures that safe controls will be used only when necessary and not unduly impact planner performance, (ii) safety is defined intrinsically in the BRS computation whereas safety using FRS-based approaches depends on whether the robot's planned trajectory intersect the FRS, and (iii) it provides a computational handle on safe controls which can be evaluated at high operating frequencies to react to split-second threats.

Moreover, we will compute the BRS using Hamilton-Jacobi (HJ) reachability analysis, a particular approach to computing reachable sets. There are many existing approaches \citep{GreenstreetMitchell1998,KurzhanskiVaraiya2000,FrehseLeGuernicEtAl2011,AlthoffKrogh2014,MajumdarVasudevanEtAl2014}, but there is always a trade-off between modeling assumptions, scalability, and representation fidelity (i.e., whether the method computes over- or under-approximations of the reachable set). Compared to alternatives approaches, HJ reachability is the most computationally expensive, but it is able to compute the BRS exactly\footnote{With precision dependent on parameters of the numerical solver, e.g., discretization choices in mesh size/time step.} for any general nonlinear dynamics with control and disturbance inputs because it essentially uses a brute force computation via dynamic programming.
As a result, the BRS solution to the HJ reachability problem represents a constructive proof of the existence of a safety-preserving control policy (i.e., a safety certificate) from states outside the BRS.
Despite the apparent computational drawbacks, we note that the BRS may be computed offline, and requires only a near-instant lookup during runtime, allowing it to be used in controllers or planners that run at a very high operational frequency.

\begin{figure*}[t]
    \centering
    \includegraphics[width=0.82\textwidth]{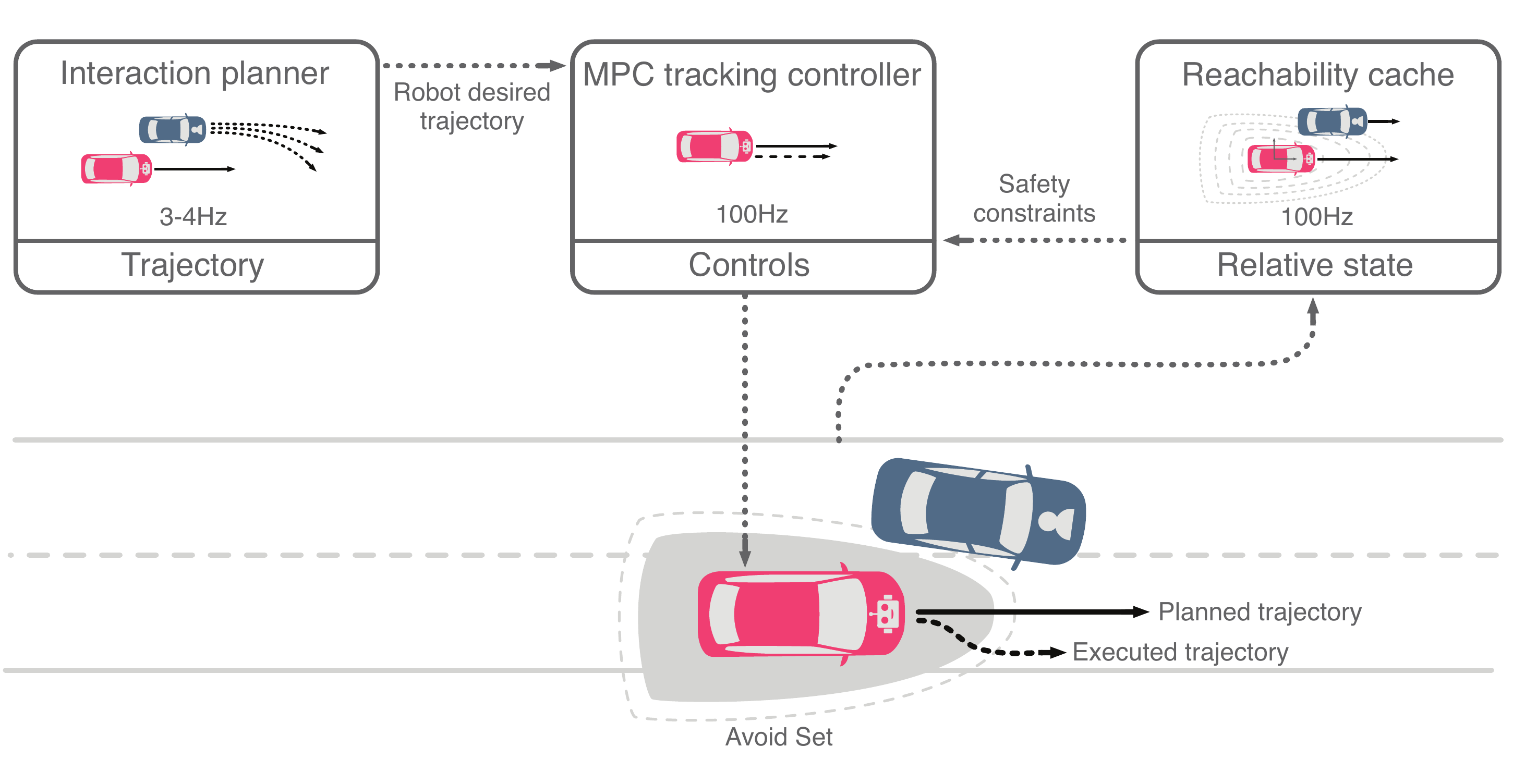}
    \caption{Decision-making and control stack for human-robot pairwise vehicle interactions.
    Our contribution in this work is the integration of safety-ensuring control constraints, derived from a HJ backward reachable set computed and cached offline, into a model predictive controller's tracking optimization problem. A high-level interaction planner produces nominal trajectories for the robot car and the low-level safe tracking controller executes controls that minimally deviate from the planner's choice if the vehicles approach the set of unsafe relative states.
    }
    \label{fig:schematic}
\end{figure*}

\subsection{Hamilton-Jacobi Backward Reachability Analysis}

We briefly review relevant HJ backward reachability definitions for the remainder of this section; see \cite{ChenTomlin2018} for a more in-depth treatment. HJ reachability casts the reachability problem as an optimal control problem and thus computing the reachable set is equivalent to solving the Hamilton-Jacobi-Isaacs (HJI) partial differential equation (PDE).

The general HJ reachability formulation is as follows. Let the system dynamics be given by $\dot{x} = f(x, u, d)$ where $x\in\mathbb{R}^n$ is the state, $u\in\mathcal{U} \subset \mathbb{R}^m$ is the control, and $d\in\mathcal{D} \subset\mathbb{R}^p$ is the disturbance. For example, $x$ could be the state of a robot, $u$ be the robot's controls, $d$ be environmental disturbances such as wind, and $f$ be the dynamics of the robot. The system dynamics $f:\mathbb{R}^n \times \mathcal{U} \times \mathcal{D} \rightarrow \mathbb{R}^n$ are assumed to be uniformly continuous, bounded, and Lipschitz continuous in $x$ for a fixed $u$ and $d$. Let $\mathcal{T} \subseteq \mathbb{R}^n$ be the target set that the system wants to avoid at the end of a time horizon $|t|$ (note that $t<0$ when propagating backwards in time). For collision avoidance, $\mathcal{T}$ typically represents the set of states that are in collision with an obstacle. For brevity, the following description of HJ backward reachability will use notation relevant to the rest of the paper, that is, we are interested in collision avoidance for human-robot interactions. Despite the notation, the following theory is still applicable to general systems $\dot{x} = f(x, u, d)$ outside the domain of human-robot interactions.

In the context of human-robot interactions, $f(\cdot)$ describes the \emph{relative dynamics} between the human and the robot (denoted by $f_{_\rel}(\cdot)$ where the subscript $\rel$ indicates the relative human-robot system), $u$ corresponds to the robot's controls, and $d$ corresponds to the human's controls since the human actions are treated as disturbance inputs.
More concretely, let $(\xrob, \urob)$ represent the robot state and control, $(\xhum, \uhum)$ represent the human state and control, and $\Xrel$ be the relative state between the human and the robot.
Thus the relative dynamics of the robot and human are given by $\dot{x}_{_\rel} = \frel(\Xrel, \urob, \uhum)$, and $\mathcal{T}$ represents the set of relative states corresponding to when the human and robot are in collision.
The formal definition of the BRS, denoted by $\mathcal{A}(t)$, for the human-robot relative system is
\begin{equation}
\begin{split}
    &\mathcal{A}(t)  := \lbrace  \bar{x}_{_\rel} \in \mathbb{R}^n: \exists \uhum(\cdot), \forall \urob (\cdot), \exists s\in[t, 0], \\
    & \Xrel(t) = \bar{x}_{_\rel}\wedge\dot{x}_{_\rel} = f_{_\rel}(\Xrel, \urob, \uhum) \wedge \Xrel(s) \in \mathcal{T} \rbrace.
\end{split}
    \label{eqn: backward reachable tube}
\end{equation}
$\mathcal{A}(t)$ represents the set of ``avoid states'' at time $t$ from which if the human followed an adversarial \emph{policy} $\uhum(\cdot)$, any robot \emph{policy} $\urob(\cdot)$ would lead to the relative state trajectory $\Xrel(\cdot)$ being inside $\mathcal{T}$ within a time horizon $|t|$.
Assuming optimal (i.e., adversarial) human actions, $\mathcal{A}(t)$ can be computed by defining a value function $V(t, \Xrel)$ which obeys the HJI PDE \citep{MitchellBayenEtAl2005,FisacChenEtAl2015}; the solution $V(t, \Xrel)$ gives the BRS as its zero sublevel set:
\[
\mathcal{A}(t) = \lbrace \Xrel : V(t,\Xrel) \leq 0\rbrace.
\]
The HJI PDE is solved starting from the boundary condition $V(0, \Xrel)$, the sign of which reflects set membership of $\Xrel$ in $\mathcal{T}$.\footnote{See Section~\ref{sec:insights} for discussion of specific choices of $V(0, \Xrel)$.} Thus computing the BRS is equivalent to solving the HJI PDE with boundary condition $V(0,\Xrel)$; we cache the solution $V(t,\Xrel)$ to be used online as a look-up table.

For the case of the vehicle-vehicle interactions investigated in this work, when the control and maximum velocity capabilities of the human car are no greater than those of the robot car, one can take the limit $t\rightarrow-\infty$ and obtain the infinite time horizon BRS $\mathcal{A}_\infty$ with corresponding value function $V_\infty(\Xrel)$.\footnote{For ease of notation going forward we will often write $V := V_\infty$.}
Intuitively, this prescribed parity in control authority ensures that if the human and robot start sufficiently far apart, then the human will never be able to ``catch'' the robot. This holds even if the human car may have transient maneuverability advantages over the robot as we assume later in this work. That is, we expect that the BRS will not encompass the entire state space as $t\rightarrow-\infty$, and in practice we compute the BRS over a sufficiently large finite time horizon to the point where it appears that the BRS has converged. We recognize that we make a strong assumption on the human and robot's control authority in enabling this computation. In reality, the human and robot car may have very different control authority (e.g., different engines resulting in different acceleration and velocity capabilities) and the infinite BRS may not be bounded. In such cases, practitioners may compute the BRS over a horizon suitable for the interaction where guarantees afforded by HJ reachability only hold over that time horizon.
Illustrative slices of the value function and the BRS for the vehicle-vehicle relative dynamics with equal control and velocity capabilities considered in this work are shown in Figure~\ref{fig:collision_avoid_set} while Figure \ref{fig:relative frame} illustrates the human-robot relative frame used to define the relative dynamics involved in the BRS computation (the mathematical details are given in Section \ref{sec: relative dynamics}). In Figure \ref{fig:collision_avoid_set} (left), the pear-shaped BRS stems from the fact the robot car can swerve its front more rapidly than its rear to avoid collision. In Figure \ref{fig:collision_avoid_set} (middle), if the robot car is traveling faster, it is unsafe for the robot car to be in the region behind the human car because a collision may be unavoidable if the human brakes abruptly. In Figure \ref{fig:collision_avoid_set} (right), if the human car is traveling faster, in this case at an angle, it is unsafe for the robot to be in the region in front of the human car because the robot car may not be able to maneuver out of the way before the human catches up.

We can compute the optimal robot collision avoidance control
\begin{equation}
u_{_\mathrm{R}}^* = \arg \max_{\urob} \min_{\uhum}\nabla V(\Xrel) ^T \frel(\Xrel, \urob, \uhum)
 \label{eqn: optimal control}
\end{equation}
which offers the greatest increase in $V(\Xrel)$ assuming optimal (worst-case) actions by the human.
For general nonlinear systems, computing the optimal collision avoidance control in Equation~\eqref{eqn: optimal control} may be nontrivial as there could be multiple local maxima. For control/disturbance affine systems, however, the solutions to the optimal control/disturbance are bang-bang.

Recall from Equation \eqref{eqn: backward reachable tube} that when the system is outside of the avoid set $\mathcal{A}(t)$ (i.e., $V>0$), there exists a robot policy (e.g., Equation \eqref{eqn: optimal control}) that keeps the robot safe over a time horizon $|t|$ regardless of any (including adversarial) policy taken by the human.
Previous applications of HJI solutions switch to the optimal control (Equation \eqref{eqn: optimal control}) when near the boundary of the BRS, i.e., when safety is nearly violated \citep{FisacAkametaluEtAl2018,BajcsyBansalEtAl2019}
This reflects the goal of HJ reachability-based safety which is to ensure that the system stays outside of the avoid set $\mathcal{A}(t)$, that is, the value function should always stay positive.

In an interactive scenario where, for example, we may want to let a robot planner convey intent by nudging towards the human car to the extent that is safe, we prefer a less extreme control strategy. In the next section, we describe in detail how to infuse reachability-based safety assurance within a multi-tiered control framework that consists of different planning objectives.

\section{Control Stack Architecture}\label{sec:approach}

In this section, we propose using a safety-preserving HJI controller, rather than switching to the optimal HJI controller defined in Equation \eqref{eqn: optimal control}, and describe how to incorporate it within an existing control stack---a high-level planner feeding desired trajectories to a low-level tracking controller---to enable safe human-robot interactions that minimally impinges on the high-level planning performance objective. The control stack architecture is illustrated in Figure \ref{fig:schematic}.
The proposed control stack is applicable to general human-robot interactions (e.g., an autonomous car interacting with a human-driven car, an autonomous manipulator arm working alongside a human, wheeled mobile robots navigating a crowded sidewalk). However, in this work, we focus on the traffic-weaving scenario, wherein two cars initially side-by-side must swap lanes in a limited amount of time and distance, because it is a representative interactive scenario that encapsulates many challenging characteristics inherent to human-robot interaction. Successful and smooth traffic-weaves rely on action anticipation, intent prediction, and proactive behavior from each vehicle, and ensuring safety is critical because collisions may lead to life-threatening injury.

\subsection{Interaction Planner}
Our proposed control framework is agnostic to the interaction planner used. The only assumption for the planner is that it outputs desired trajectories (which presumably reflect high-performance goal-oriented nominal behavior) for the robot car to track.
In general, high-level planners often optimize objectives that weigh safety considerations (e.g., distance between cars) relative to other concerns (e.g., control effort), and typical to human-robot interactive scenarios, they may reason anticipatively with respect to a \emph{probabilistic} interaction dynamics model. That is, although the planner is encouraged to select safer plans, safety is not enforced as a deterministic constraint at the planning level.

In this work, we use the traffic-weaving interaction planner from \cite{SchmerlingLeungEtAl2018}. It uses a predictive model of future human behavior to select desired trajectories for the robot car to follow, updated at $\sim$3Hz. We extend this work by using a hindsight optimization policy \citep{YoonFernEtAl2008} instead of the limited-lookahead action policy in order to encourage more information-seeking actions from the robot.

\subsection{Tracking Controller}
Given a desired trajectory from the interaction planner, the tracking controller computes optimal controls to track the desired trajectory. We assume that the outputs of the low-level tracking controller are directly usable by the robot's actuators (e.g., steering and longitudinal force commands for a vehicle, torque commands for each joint on a manipulator arm). As such, the tracking controller typically uses a more accurate dynamics model and operates at a much higher frequency ($\sim 100$Hz) than the interaction planner, often at the expense of environmental awareness.
To ensure safety with respect to a dynamic obstacle (i.e., a human-driven vehicle) we incorporate additional constraints computed from HJ reachability analysis into the tracking optimization problem.
These constraints are designed to ensure that at each control step the robot car does not enter an unsafe set of relative states that may lead to collision.

In this work, we are concerned with vehicle trajectory tracking. We adapt the real-time model predictive control (MPC) tracking controller from \cite{BrownFunkeEtAl2017} by modifying it to include an additional invariant set constraint detailed in the next section.
This MPC tracking controller, operating at 100Hz, computes optimal controls to track a desired trajectory by solving an optimization problem at each iteration.
The optimization problem is based on a single track vehicle model (also known as the bicycle model) and incorporates friction and stability control constraints (in addition to control and state constraints) while minimizing a combination of tracking error and control derivatives.
A more in-depth treatment of this combined MPC and HJI controller is given in Section \ref{sec:mpc}.

\subsection{Safety-Preserving HJI Control}

Rather than switching to the optimal avoidance controller defined in Equation \eqref{eqn: optimal control} when nearing safety violation, we propose adding containment in the \emph{set of safety-preserving controls} as an additional constraint to the low-level MPC tracking controller.
The set of safety-preserving controls
\begin{equation}
    \mathcal{U}_{_\mathrm{R}}(\Xrel) = \{\urob: \min_{\uhum} \nabla V(\Xrel) ^T \frel(\Xrel, \urob, \uhum) \ge 0\}
    \label{eq:safe preserve}
\end{equation}
represent the set of robot controls that ensure the value function is nondecreasing.
This constraint can be computed online since the BRS is computed offline and the value function $V$ and its gradient $\nabla V$ are cached.
Online we employ a safety buffer $\epsilon>0$ so that when the condition $V(\Xrel) \leq \epsilon$ holds, indicating that the robot is nearing safety violation, we add the constraint $\urob\in \mathcal{U}_{_\mathrm{R}}(\Xrel)$ to the list of tracking controller constraints.
By adding this additional safety-preserving constraint when near safety violation, the MPC tracking controller selects control actions that prevent the robot from further violating the safety threshold while simultaneously optimizing for tracking performance and obeying additional constraints. This results in a \emph{minimally interventional safety controller}---the MPC tracking controller will only minimally deviate from the desired trajectory to the extent necessary to maintain safety for the robot car.

For the traffic-weaving scenario investigated in this paper, the MPC problem for vehicle trajectory tracking \citep{BrownFunkeEtAl2017} is formulated as a quadratic program (QP) (to enable fast solve time amenable to a 100Hz operating frequency) and hence requires the constraints to be linear. As such, we instead apply the constraint $\urob \in \widetilde{\mathcal{U}}_R(\Xrel)$ where
\begin{equation}
\widetilde{\mathcal{U}}_{_\mathrm{R}}(\Xrel)= \lbrace \urob:  M_{\HJI}^T \urob + b_{\HJI}\geq 0 \rbrace
\label{eqn:half-plane constraint}
\end{equation}
is a linearized approximation of $\mathcal{U}_{_\mathrm{R}}(\Xrel)$.
Specifically, for the current relative state $\tilde{x}_{_\rel}$, current robot control $\tilde{u}_{_\mathrm{R}}$, and optimal, i.e., worst-case, human action defined analogously to Equation~\eqref{eqn: optimal control} $\uhum^*$, the terms in the linearization are
$$
M_{\HJI} = \frac{\partial }{\partial \urob}\left( \nabla V  ^T \frel(\Xrel, \urob, \uhum)\right) \Bigr\rvert_{(\tilde{x}_{_\rel}, \uhum^*, \tilde{u}_{_\mathrm{R}})}
$$

$$b_{\HJI} = \nabla V ^T \frel(\tilde{x}_{_\rel}, \uhum^*, \tilde{u}_{_\mathrm{R}}) - M_{\HJI}^T \tilde{u}_{_\mathrm{R}}.$$
In general, $\mathcal{U}_{_\mathrm{R}}(\Xrel)$ may not be a half-space, leading to the linear approximation $\widetilde{\mathcal{U}}_{_\mathrm{R}}(\Xrel)$ including controls outside of $\mathcal{U}_{_\mathrm{R}}(\Xrel)$. However, since we bound the change in the control inputs across each time step, feasible controls remain close to the linearization point where the approximation error is small.

\section{Dynamics}
\label{sec:dynamics}
In this section, we detail the vehicle dynamics model used to model the robot and human car, the relative dynamics model between the human and the robot necessary for computing the BRS, and the tracking dynamics used for the MPC tracking controller. We use a six-state nonlinear single track model to describe the robot car's dynamics and assume the human car obeys a four-state dynamically extended nonlinear unicycle model with longitudinal acceleration and yaw rate as control inputs. As a result, the relative dynamics model has seven states. This represents a compromise between model fidelity and the number of state dimensions in the relative dynamics; increasing the former reduces the amount of model mismatch with the real system while reducing the latter is essential since solving the HJI PDE suffers greatly from the curse of dimensionality. The computation becomes notoriously expensive past five or more state dimensions without compromising on grid discretization or employing some decoupling strategy.

\subsection{Robot Vehicle Dynamics}

\begin{figure}[tb]
    \centering
    \includegraphics[width=0.45\textwidth]{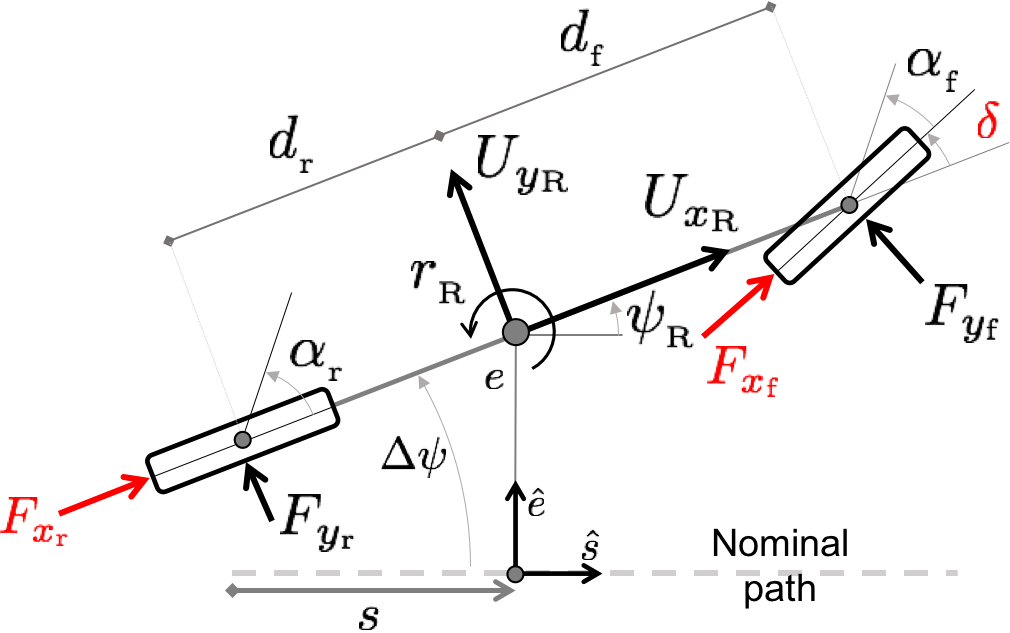}
    \caption{Schematic of the single track model (bicycle model) tracking a path. This is the dynamics model used to describe the robot vehicle.}
    \label{fig:bicycle model diagram}
\end{figure}

The robot car (denoted by subscript $\mathrm{R}$) will be modeled using the single track vehicle model illustrated in Figure \ref{fig:bicycle model diagram}. Let $(\xr,\yr)$ be the position of the robot car's center of mass defined in an inertial reference frame and $\psir$ be the yaw angle (heading) of the robot car relative to the horizontal axis. $\Uxr$ and $\Uyr$ are the velocities in the robot car's body frame, and $\rr$ is the yaw rate. The state for the single track model is $\xrob = \begin{bmatrix} \xr & \yr & \psir & \Uxr & \Uyr & \rr \end{bmatrix}^T$.
The control input $\urob = \begin{bmatrix} \delta & \Fx\\ \end{bmatrix}^T$ consists of the steering command $\delta$ and longitudinal tire force $\Fx$ which is distributed between the front and rear tires $\Fx = \Fxf + \Fxr$ via a fixed mapping.
Assuming a quadratic model of longitudinal drag force $(F_{x_\mathrm{drag}} = -C_{d_0} - C_{d_1}\Uxr - C_{d_2}\Uxr^2)$ and for vehicle mass $(m)$ and moment of inertia $(I_{zz})$, and the distances from the center of mass to the front and rear axles ($d_\mathrm{f}$, $d_\mathrm{r}$), the equations of motion for the robot car are

\begin{equation}
\dxrob = \begin{bmatrix}
\Uxr \cos{\psir} - \Uyr \sin{\psir} \\
\Uxr \sin{\psir} + \Uyr \cos{\psir}\\
r\\
\frac{1}{m} (\Fxf \cos \delta - \Fyf \sin \delta + \Fxr + F_{x_\mathrm{drag}}) + \rr \Uyr\\
\frac{1}{m}(\Fyf \cos \delta + \Fyr + \Fxf \sin \delta) - \rr\Uxr\\
\frac{1}{I_{zz}}(d_{_\mathrm{f}}\Fyf \cos \delta + d_{_\mathrm{f}} \Fxf \sin \delta - d_{_\mathrm{r}}\Fyr)\\
\end{bmatrix}
\label{eqn:bicycle model}
\end{equation}

The controls are assumed to be limited by the steering system, friction limits, and power capacity of the vehicle.
Using the brush coupled tire model by \cite{Pacejka2002}, the lateral tire force $\Fyf$ and $\Fyr$ at the front and rear tires is a function of slip angle ($\alpha_\mathrm{f}, \,\alpha_\mathrm{r}$), tire cornering stiffness ($C_{\alpha_\mathrm{f}}, \,C_{\alpha_\mathrm{r}}$), longitudinal tire forces ($\Fxf,\, \Fxr$), coefficient of friction ($\mu$), and normal tire forces ($\Fxf, \, \Fzr$). As such, the lateral tire force for either the front or rear tires (denoted by $i \in \{\mathrm{f}, \mathrm{r}\}$) is
\begin{equation}
F_{y_i} = \begin{cases}
0 \quad &\text{if} \; F_{x_i} > \mu F_{z_i}\\
-C_{\alpha_i} \tan \alpha_i \left( 1 - \gamma + \frac{1}{3} \gamma^2\right)  \quad &\text{if} \; \gamma < 1\\
-F_{y_{\max}}\mathrm{sign}(\tan \alpha_i)  \quad &\text{if} \; \gamma \geq 1\\
\end{cases}
\label{eqn:fiala tire model}
\end{equation}
where $\gamma = \left|\frac{C_{\alpha_i} \tan \alpha_i }{3 F_{y_{\max}}} \right|$ and $F_{y_{\max}} = \sqrt{\mu^2 F_{z_i}^2 - F_{x_i}^2}$.
The slip angles and normal forces (accounting for weight transfer due to $F_{x_i}$) for the front and rear tires can be computed using the following equations

\begin{align*}
\alpha_\mathrm{f} = \tan^{-1}\frac{\Uyr +  d_{_\mathrm{f}}\rr}{\Uxr} - \delta,& \qquad \alpha_\mathrm{r} = \tan^{-1}\frac{\Uyr - d_{_\mathrm{r}}\rr}{\Uxr},\\
\Fzf = \frac{mgd_{_\mathrm{r}} - h \widetilde{F}_x}{L},& \qquad \Fzr = \frac{mg d_{_\mathrm{f}} + h \widetilde{F}_x}{L},
\end{align*}
where $h$ is the distance from the center of mass to the ground, $L =  d_{_\mathrm{f}} +  d_{_\mathrm{r}}$, and $\widetilde{F}_x = \Fxf\cos\delta - \Fyf\sin\delta + \Fxr$ is the total longitudinal force in the vehicle's body frame.

\subsection{Human Vehicle Dynamics}

\begin{figure}[tb]
    \centering
    \includegraphics[width=0.25\textwidth]{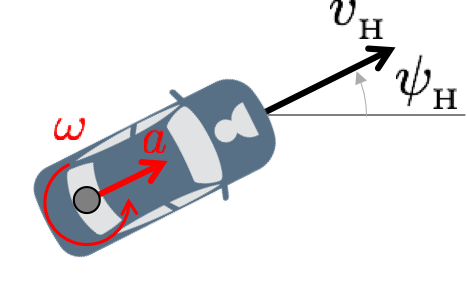}
    \caption{Schematic of the dynamically-extended unicycle model. This is the dynamics model used to describe the human vehicle.}
    \label{fig:human car}
\end{figure}

The human car (denoted by subscript $\mathrm{H}$) will be modeled using the dynamically extended unicycle model illustrated in Figure \ref{fig:human car}. Let $(\xh,\yh)$ be the position of the center of the human car's rear axle defined in an inertial reference frame and $\psih$ be the yaw angle (heading) of the human car relative to the horizontal axis. The velocity of the human car in the vehicle frame is $\vh$. The state for the dynamically extended unicycle model is
$\xhum = \begin{bmatrix}\xh & \yh & \psih & \vh \end{bmatrix}^T$.
The control input $\uhum = \begin{bmatrix} \omega & a\end{bmatrix}^T$ consists of the yaw rate $\omega$ and longitudinal acceleration $a$. The control limits of the human car are chosen such that the robot and human car share the same power, steering, and friction limits.
The equations of motion for the human car are
\begin{equation}
\dot{x}_{_\mathrm{H}} =
\begin{bmatrix}
\vh \cos{\psih}\\
\vh \sin{\psih} \\
\omega \\
a
\end{bmatrix}.
\label{eqn:dynamically extended simple car}
\end{equation}

Due to its simpler dynamics representation, the human car has a transient advantage in control authority over the robot car (it may change its path curvature discontinuously, while the robot may not), but by equating the steady-state control limits we ensure that the infinite time horizon BRS computation converges.

\subsection{Relative Dynamics}\label{sec: relative dynamics}
The relative state (denoted by subscript $\rel$) between the robot car and human car is defined with respect to a coordinate system centered on and aligned with the robot car's vehicle frame. We define the relative position $(\xrel, \yrel)$ as
\begin{align*}
\begin{bmatrix}
\xrel\\ \yrel
\end{bmatrix} = \begin{bmatrix}
\cos\psir & \sin\psir \\ -\sin\psir & \cos\psir
\end{bmatrix}\begin{bmatrix}
\xh - \xr \\ \yh - \yr
\end{bmatrix},
\end{align*} and the relative heading $\psirel$ as $\psirel = \psih - \psir$.

Since the velocity states are defined with respect to the vehicle frame, we cannot define analogous relative velocity states and must include the individual velocity states of each vehicle. As such, the relative state for the human-robot vehicle system is $\Xrel = \begin{bmatrix} \xrel & \yrel & \psirel & \Uxr & \Uyr & \vh & \rr \end{bmatrix}^T$. In the language of HJ reachability, the disturbance input of the system is the human car's control $d = \uhum = \begin{bmatrix} \omega & a
\end{bmatrix}^T$ and the control input is the robot car's control $u = \urob = \begin{bmatrix} \delta & \Fx
\end{bmatrix}^T$.
Combining Equations \eqref{eqn:bicycle model} and \eqref{eqn:dynamically extended simple car}, the equations of motion for the relative system are

\begin{align}
\dot{x}_{_\rel} = \begin{bmatrix}
\vh \cos \psirel - \Uxr + \yrel \rr\\
\vh \sin \psirel - \Uyr - \xrel \rr\\
\omega - \rr\\
\frac{1}{m} (\Fxf \cos \delta - \Fyf \sin \delta + \Fxr + F_{x_\mathrm{drag}}) + \rr \Uyr\\
\frac{1}{m}(\Fyf \cos \delta + \Fyr + \Fxf \sin \delta) - \rr\Uxr\\
a\\
\frac{1}{I_{zz}}(d_{_\mathrm{f}}\Fyf \cos \delta + d_{_\mathrm{f}} \Fxf \sin \delta - d_{_\mathrm{r}}\Fyr)\\
\end{bmatrix}.
\label{eqn:relative dynamics}
\end{align}

\subsection{Tracking Dynamics}

The tracking MPC controller relies on an error dynamics model.
Define a path (see Figure \ref{fig:bicycle model diagram}) through space where $s$ is the distance along the path and at any distance $s$, we know the path heading $\psi(s)$, the curvature $\kappa(s)$, and a coordinate system $(\hat{s}, \hat{e})$ tangential and normal to the path. Given the position and heading of the robot car, we can project the car to the closest point on the path. Let the lateral error $e$ be the lateral distance along direction $\hat{e}$ and $\Delta\psi = \psir - \psi_\mathrm{path}$ be the robot car's heading relative to the path computed from this closest point.
Using this projection and the robot car's dynamics from Equation \eqref{eqn:bicycle model}, we can compute the error dynamics relative to the desired path. Using the same notation defined previously in Equation \eqref{eqn:bicycle model}, the state for the robot car tracking a desired path is $\hat{x} = \begin{bmatrix}
s & \Uxr & \Uyr & \rr & \Delta\psi & e \end{bmatrix}^T$ and the controls $u = \urob = \begin{bmatrix} \delta & \Fx
\end{bmatrix}^T$. The tracking dynamics are

\begin{align}
\dot{\hat{x}} = \begin{bmatrix}
\Uxr \cos\Delta\psi - \Uyr \sin\Delta\psi\\
\frac{1}{m} (\Fxf \cos \delta - \Fyf \sin \delta + \Fxr + F_{x_{d}}) + \rr \Uyr\\
\frac{1}{m}(\Fyf \cos \delta + \Fyr + \Fxf \sin \delta) - \rr\Uxr\\
\previouslyrevised{\frac{1}{I_{zz}}(d_{_\mathrm{f}}\Fyf \cos \delta + d_{_\mathrm{f}} \Fxf \sin \delta - d_{_\mathrm{r}}\Fyr)}\\
r - \left(\Uxr \cos\Delta\psi - \Uyr \sin\Delta\psi\right)\kappa(s)\\
\Uxr\sin\Delta\psi + \Uyr\cos\Delta\psi
\end{bmatrix}.
\label{eqn:tracking dynamics}
\end{align}

\section{The MPC+HJI Trajectory Tracking Controller}
\label{sec:mpc}
In this section, we detail the optimization problem used in the MPC+HJI tracking controller, describe how this problem can be adapted to accommodate collision avoidance for static obstacles, and provide numerical details about the BRS computation used in this formulation.
Central to an MPC controller is an optimization problem; at each time step, the controller solves an optimization problem to find an optimal control sequence, passes the first control input to the actuator, and then repeats this process. To be amenable to real-time applications, the optimization problem requires a fast solve time ($\sim 0.01$s). In this work, the MPC tracking problem is formulated as a convex optimization problem, namely a QP, enabling the use of efficient solvers \citep{MattingleyBoyd2012,StellatoBanjacEtAl2017} which are capable of solving the QP within the tight operating frequency.

\subsection{Optimization Problem}
Both the trajectory tracking objective and safety-preserving control constraint rely on optimizing over the robot steering and longitudinal force inputs simultaneously.
Let $q_k = \begin{bmatrix} \Delta s_k & U_{x_\mathrm{R},k} & U_{y_\mathrm{R},k} & r_{_\mathrm{R},k} & \Delta \psi_k & e_k\end{bmatrix}^T$ be the state of the robot car with respect to a nominal trajectory at discrete time step $k$. $\Delta s_k$, $e_k$ and $\Delta \psi_k$ denote longitudinal, lateral, and heading error; $U_{x_\mathrm{R},k}$, $U_{y_\mathrm{R},k}$, and $r_{_\mathrm{R},k}$ are body-frame longitudinal and lateral velocity, and yaw rate respectively as defined in Figure \ref{fig:bicycle model diagram}. Let $u_k=\begin{bmatrix} \delta_k & F_{x,k}\end{bmatrix}^T$ be the controls at step $k$ and let $A_k q_k + B_k^- u_k + B_k^+ u_{k+1} + c_k = q_{k+1}$ denote linearized first-order-hold dynamics of Equation \eqref{eqn:tracking dynamics}.
We adopt the varying time steps method ($N_\mathrm{short}$ time steps of size $\Delta t_\mathrm{short}$ and $N_\mathrm{long}$ time steps of size $\Delta t_\mathrm{long}$) and stable handling envelope constraint from \cite{BrownFunkeEtAl2017} (expressed as $H_k$ and $G_k$ in the problem formulation below). To ensure the existence of a feasible solution, we use slack variables $\sigma_{\beta,k}$, $\sigma_{r,k}$, and $\sigma_{\HJI,k}$ on the stability and HJI constraints. The HJI reachability constraint $M_{\HJI} u_k + b_{\HJI} \geq -\sigma_{\HJI}$ is activated only when $V(\Xrel) \leq \epsilon$.
Although HJI theory suggests that applying this constraint on the next action alone is sufficient, we apply it over the next $N_\HJI=3$ timesteps (30ms lookahead) to account for the approximations inherent in our QP formulation. The MPC tracking problem is a quadratic program of the form
\begin{equation}
\begin{aligned}
\minimize_{q, u, \sigma, \sigma_{\HJI}, \Delta\delta, \Delta F_x}  & \;\;\sum_{k=1}^T \bigg(\Delta s_k ^T Q_{\Delta s} \Delta s_k + \Delta \psi_k ^T Q_{\Delta \psi} \Delta \psi_k + \\
& e_k ^T Q_e e_k  +  \left(\frac{\Delta \delta_k}{\Delta t_k}\right) ^T R_{\Delta  \delta}  \left(\frac{\Delta \delta_k}{\Delta t_k}\right)+ \\
& \left(\frac{\Delta F_{x,k}}{\Delta t_k}\right)^T R_{\Delta \Fx} \left(\frac{\Delta F_{x,k}}{\Delta t_k}\right) +\\
& W_{\beta} \sigma_{\beta,k} + W_r\sigma_{r,k} + W_{\HJI}\sigma_{\HJI,k}\bigg)\Delta t_k\\
\text{subject to} \quad & q_1 = q_\mathrm{curr}, \quad u_1=u_\mathrm{curr},\\
                        & \delta_{k+1} - \delta_k = \Delta \delta_k,\\
                        & \dot{\delta}_{\min} \Delta t_k \leq \Delta \delta_k \leq \dot{\delta}_{\max} \Delta t_k,\\
                        & \delta_{\min} \leq \delta_{k} \leq \delta_{\max},\\
                        & F_{x,k+1} - F_{x,k} = \Delta F_{x,k},\\
                        & F_{x,\min} \leq F_{x,k} \leq F_{x,\max},\\
                        & U_{x,\min} \leq U_{x_\mathrm{R},k} \leq U_{x,\max},\\
                        & \sigma_{\beta,k} \geq 0, \quad \sigma_{r,k} \geq 0,\\
                        & A_k q_k + B_k^- u_k + B_k^+ u_{k+1} + c_k = q_{k+1} ,\\
                        & H_k \begin{bmatrix} U_{y,k} \\ r_k \end{bmatrix} - G_k \leq \begin{bmatrix} \sigma_{\beta,k} \\ \sigma_{r,k} \end{bmatrix},\\
                        &\text{for } k=1,...,N_\mathrm{short} + N_\mathrm{long}\\
& \sigma_{\HJI,j} \geq 0  \;\; (\text{if } V(\Xrel) \leq \epsilon),\\
& M_{\HJI} u_j + b_{\HJI} \geq -\sigma_{\HJI} \;\; (\text{if } V(\Xrel) \leq \epsilon),\\
&\text{for } j = 1,...,N_{\HJI}.
\end{aligned}
\label{eqn:QP}
\end{equation}

The objective strives to minimize a combination of tracking error (longitudinal, lateral and angular), control rates (steering and longitudinal tire forces), and magnitude of the slack variables. The constraints ensure (i) continuity with the current and next state and control, (ii) the change in controls across each time step is bounded, (iii) the positivity of slack variables, (iv) the (linearized) dynamics are satisfied, (v) the vehicle stability constraints are satisfied, and (vi) the HJI safety-preserving half-plane constraint is satisfied when $V(\Xrel) \leq \epsilon$.

The time-discretization and linearizations (dynamics and constraints) we apply amount to an approximate variant of sequential quadratic programming (SQP). In particular, however, we solve one QP at each MPC step rather than the usual iteration until convergence. Since the tracking problems are so similar from one MPC step to the next, we find that this approach yields sufficient performance for our purposes. We interpolate along each solution trajectory to compute the linearization nodes for the QP at the next MPC step.

We use the \verb|ForwardDiff.jl| automatic differentiation (AD) package implemented in the Julia programming language \citep{RevelsLubinEtAl} to linearize the trajectory tracking dynamics as well as the HJI relative dynamics for the safety-preserving constraint.
We call the Operator Splitting Quadratic Program (OSQP) solver \citep{StellatoBanjacEtAl2017} through the \verb|Parametron.jl| modeling framework \citep{Koolencontributors2018}; this combination of software enables us to solve the following MPC optimization problem at 100Hz.
The MPC code, including optimization parameters and vehicle parameters (also included in Appendix~\ref{app:vehicle parameters} and \ref{app:trajectory parameters}), can be found here: \url{https://github.com/StanfordASL/Pigeon.jl}.

\begin{figure}[tb]
    \centering
    \includegraphics[width=0.5\textwidth]{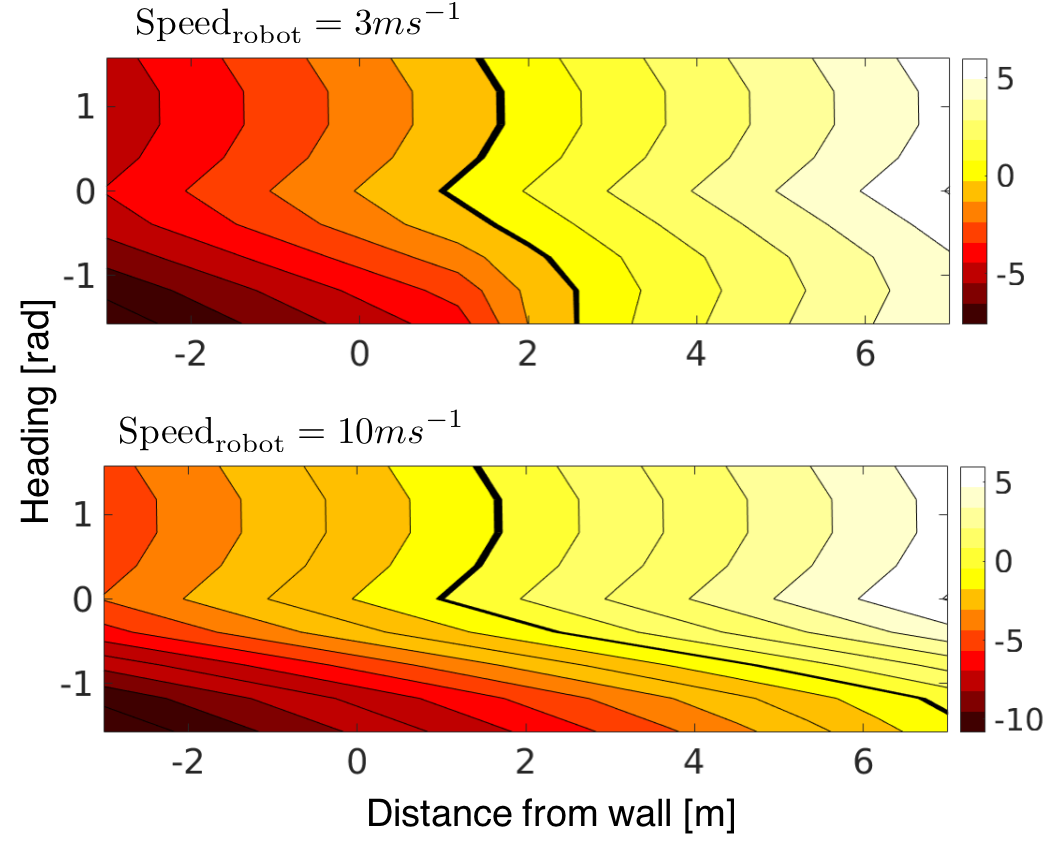}
    \caption{The BRS of the robot-wall system where zero heading corresponds to the robot car being parallel to the wall.}
    \label{fig:BRS wall}
\end{figure}

\subsection{Adding Static Obstacles}\label{sec:wall_def}
The current formulation does not prevent the robot from driving very far from its nominal trajectory (e.g., completely off the road) in order to avoid the human. In more realistic road settings, there could be environmental constraints such as a concrete road boundary.
In the same way that collision avoidance with a human-controlled vehicle is formulated as an additional constraint to the robot's MPC problem, we can add another safety constraint to account for a static wall to prevent the robot from swerving completely off the road. We consider two approaches for deriving this constraint: the first based on an additional HJI computation and giving rise to a similar control constraint applied over the first $N_\HJI$ timesteps, and the second treating the wall as a static obstacle with associated state constraints that apply over the whole duration of the MPC trajectory optimization.

For the first approach, we can compute a value function $V_{\WALL}(\xrob)$ describing the interaction between the robot and the wall by using Equation \eqref{eqn:bicycle model} without the $\xr$ state (we only care about the distance from the wall and not how far along the wall the robot is). Two slices of the robot-wall BRS projected on the $\yr-\psir$ plane are illustrated in Figure \ref{fig:BRS wall}. Intuitively, if the robot is moving towards the wall, the faster it is moving, the further away it needs to be away from the wall in order for there to exist an optimal collision avoidance control.
Note that since the wall is static, there is no disturbance input. The robot-wall HJI control constraint $M_{\WALL}\urob + b_{\WALL} \geq 0$ becomes active when $V_{\WALL}(\xrob) \leq \epsilon$. This means that it is possible for both HJI-safety constraints to be active simultaneously. We note that theoretically we could consider the robot, human, and wall simultaneously by computing the BRS for the joint system. However, naively increasing the state size without any decomposition would be computationally undesirable even offline. Thus we treat the human-robot and robot-wall systems separately, but will discuss later the impact of this design choice.

Alternatively, we can account for a static wall by adding lateral error bound constraints into the MPC tracking problem---this is the approach taken in \cite{BrownFunkeEtAl2017}. This involves always adding a left and right lateral error constraint ($e_{\min,k} \leq e_k \leq e_{\max, k}$) on each node point along the MPC trajectory such that the lateral deviation from the desired trajectory does not exceed the lateral distance to the wall. Specifically, we add the following constraints to the QP in Equation \eqref{eqn:QP}:
\[
e_k - e_{k,\min} \geq-\sigma_e,\ e_{k,\max} - e_k \geq -\sigma_e\quad\text{ for }\ k=1,...,N
\]
as well as a slack penalty, $W_e \sigma_e$ to the cost function. This approach ensures no collision with the wall over the MPC time horizon only (in contrast to HJI which is over an infinite time horizon) and in general provides higher tracking performance because the MPC controller can optimize tracking states and controls over the entire tracking trajectory.
We investigate both these approaches and provide more discussion in Section \ref{sec:insights}.

\subsection{BRS Computation}
We use the BEACLS toolkit \citep{TanabeChen2018} implemented in C++ to compute the BRS. Since HJ reachability suffers from the curse of dimensionality, this is, to the best of our knowledge, the first attempt to use HJ reachability to compute the BRS for a seven-state relative system, especially with such high modeling fidelity. We, however, do sacrifice on grid size and use a relatively coarse grid compared to other literature standards. We linearly interpolate between grid points in order to evaluate the value function and its gradient at any given state (human-robot relative state or robot-wall state).
We use a grid size of $13\times 13\times 9\times 9\times 9\times 9\times 9$ for our 7D system uniformly spaced over $(\xrel, \yrel, \psirel, \Uxr, \Uyr, \vh, \rr) \in [-15, 15] \times [-5, 5] \times [-\pi/2, \pi/2] \times [1, 12] \times [-2, 2] \times [1, 12] \times [-1, 1]$; computing the BRS with this system and discretization takes approximately 70 hours on a 3.0GHz octocore AMD Ryzen 1700 CPU.

Computing the BRS requires computing the optimal control and disturbance defined in Equation \eqref{eqn: optimal control}. For the relative dynamics model, the optimal disturbance (i.e., human actions) is a bang-bang solution since the disturbance is affine. Due to the highly nonlinear nature of the dynamics, we use a uniform grid search across $\delta$ and $\Fx$ to calculate optimal actions (i.e., robot actions).

For the robot-wall system, we compute the optimal control actions in the same fashion. We use a grid size of $21\times 9\times 9\times 9\times 9$ uniformly spaced over $(\yr, \psir, \Uxr, \Uyr, \rr) \in [-3, 7] \times [-\pi/2, \pi/2] \times [1, 12] \times [-2, 2] \times [-1, 1]$; computing the BRS with this system and discretization takes approximately 40 minutes.

\section{Results}

\begin{figure}[t]
    \centering
    \includegraphics[width=0.45\textwidth]{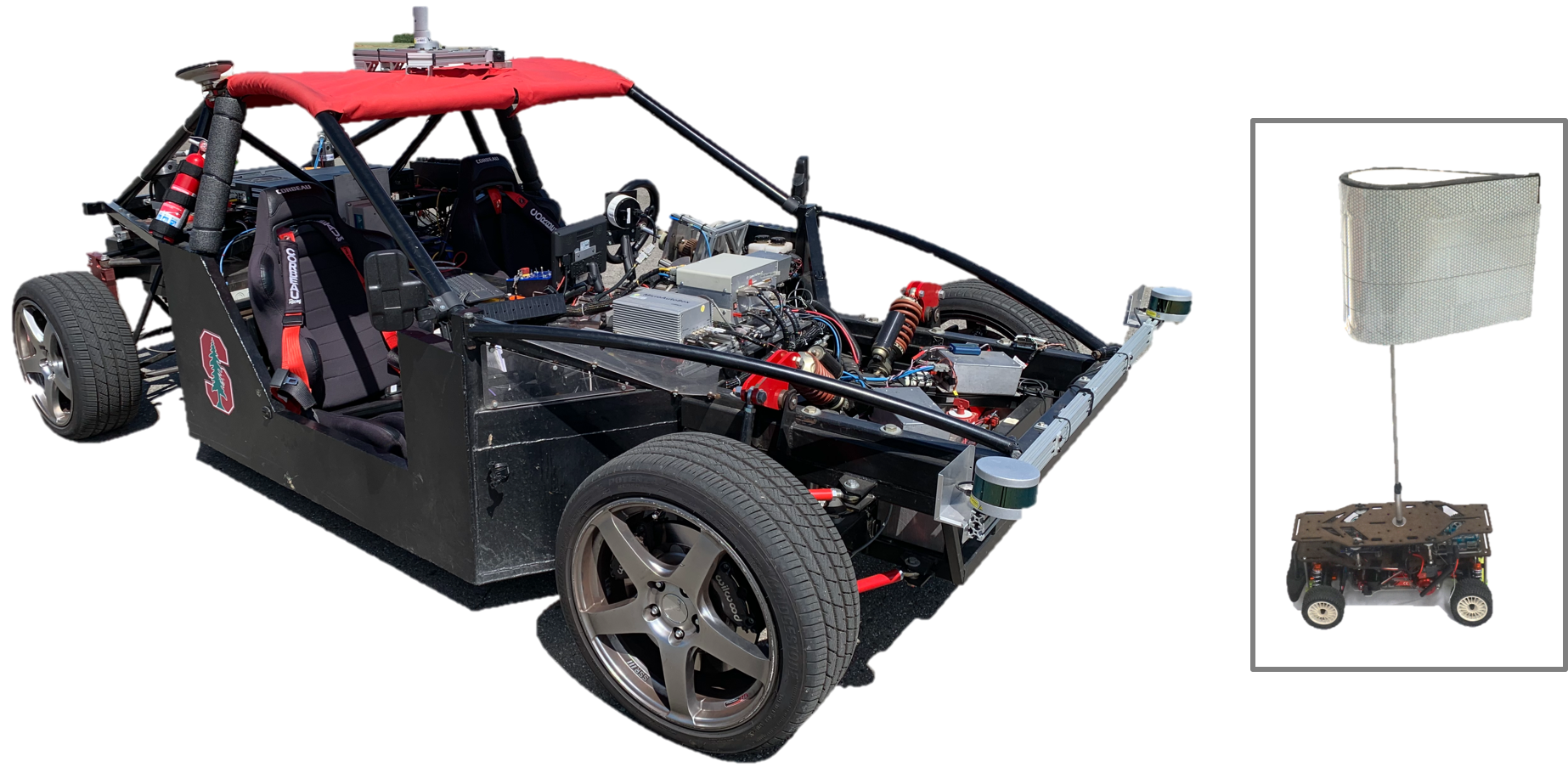}
    \caption{X1: a steer-by-wire experimental vehicle platform. The 1/10-scale RC car has a mast that is visible to the LiDARs.}
    \label{fig:X1 label}
\end{figure}

\begin{figure*}[ht]
\centering
\subfloat[Controller comparison on experimental data where the robot car uses the MPC+HJI controller.]{
    \label{fig:HJI+MPC comparison}
    \includegraphics[width=\textwidth]{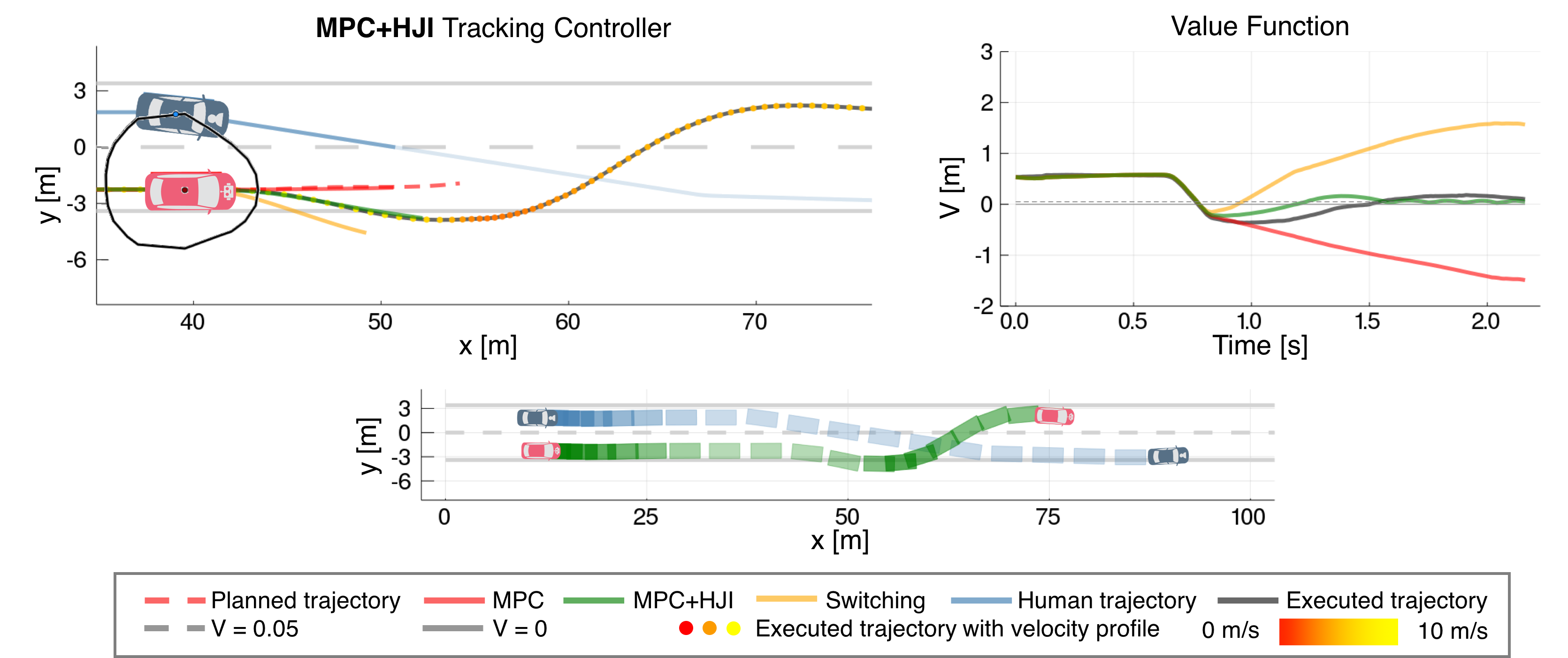} }

\subfloat[Controller comparison on experimental data where the robot car uses the switching controller.]{
    \label{fig:switching comparison}
    \includegraphics[width=\textwidth]{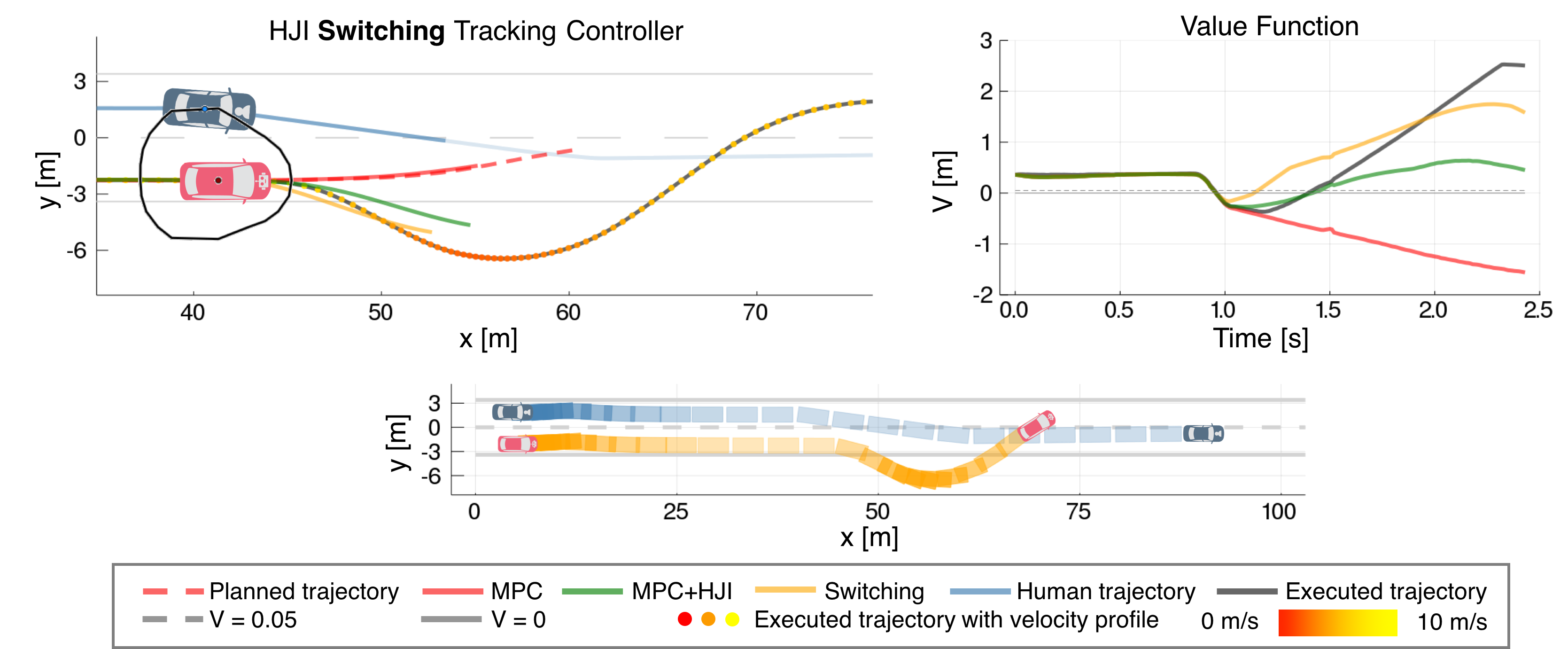}}
\caption{Controller comparison on planner trajectories from X1-virtual human-driven vehicle experiments. Left: Simulations of using the MPC, MPC+HJI, and switching controllers when $V(\Xrel(t))$ first drops below $\epsilon = 0.05$ are shown, and are compared against the executed trajectory (from experiments) and the desired trajectory (from the interaction planner). Right: The corresponding evolutions of $V(\Xrel(t))$. Bottom: Illustration of the traffic-weaving interaction. The human controlled car (blue car) carelessly drives into the path of the robot car (red car), causing the robot car to react by swerving. The transparency of the rectangles corresponds to the speed of the car (higher transparency corresponds to higher speed).
}
\label{fig:controller comparison}
\end{figure*}

\subsection{Experimental Vehicle Platform}

X1 is a flexible steer-by-wire, drive-by-wire, and brake-by-wire experimental vehicle developed by the Stanford Dynamic Design Lab (see Figure \ref{fig:X1 label}). It is equipped with three LiDARs (one 32-beam and two 16-beam), a differential GPS/INS which provides 100Hz pose estimates accurate to within a few centimeters as well as high fidelity velocity, acceleration, and yaw rate estimates.
To control X1, desired steering ($\delta$) and longitudinal tire force ($\Fxf, \Fxr$) commands are sent to the dSpace MicroAutoBox (MAB) which handles all sensor inputs except LiDAR (handled by the onboard PC) and implements all low level actuator controllers. Similarly, state information about the vehicle is obtained from the MAB.
We use the Robot Operating System (ROS) to communicate with the MAB which handles sensing and control at the hardware level; the planning/control stack described in this work is running onboard X1 on a consumer desktop PC running Ubuntu 16.04 equipped with a quadcore Intel Core i7-6700K CPU and an NVIDIA GeForce GTX 1080 GPU.
X1 parameters used in the equations of motion (Equation \eqref{eqn:bicycle model}) are listed in Appendix \ref{app:vehicle parameters}.

We also perform experiments using a LiDAR-visible 1/10-scale RC car (Figure \ref{fig:X1 label} right) as the human-driven car to investigate the robustness of our proposed control stack with perception uncertainty.

The code used to run the experiments can be found here: \url{https://github.com/StanfordASL/safe_traffic_weaving}.

\begin{figure}
\centering
\subfloat[Trade-off between total safety and average efficiency.]{
    \label{fig:tradeoff average}
    \includegraphics[width=0.45\textwidth]{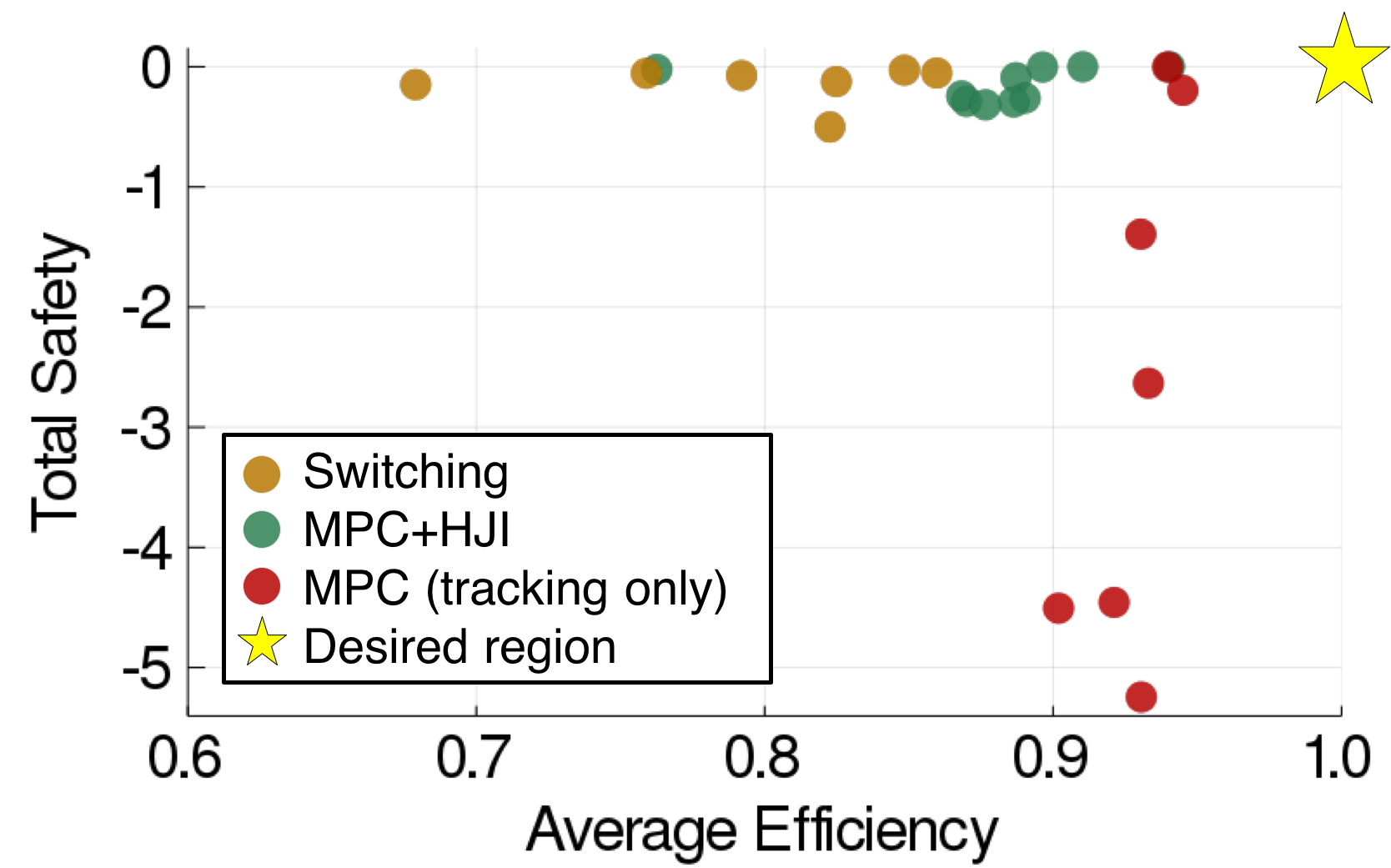} }

\subfloat[Trade-off between worst-case safety and worst-case efficiency.]{
    \label{fig:tradeoff worst-case}
    \includegraphics[width=0.45\textwidth]{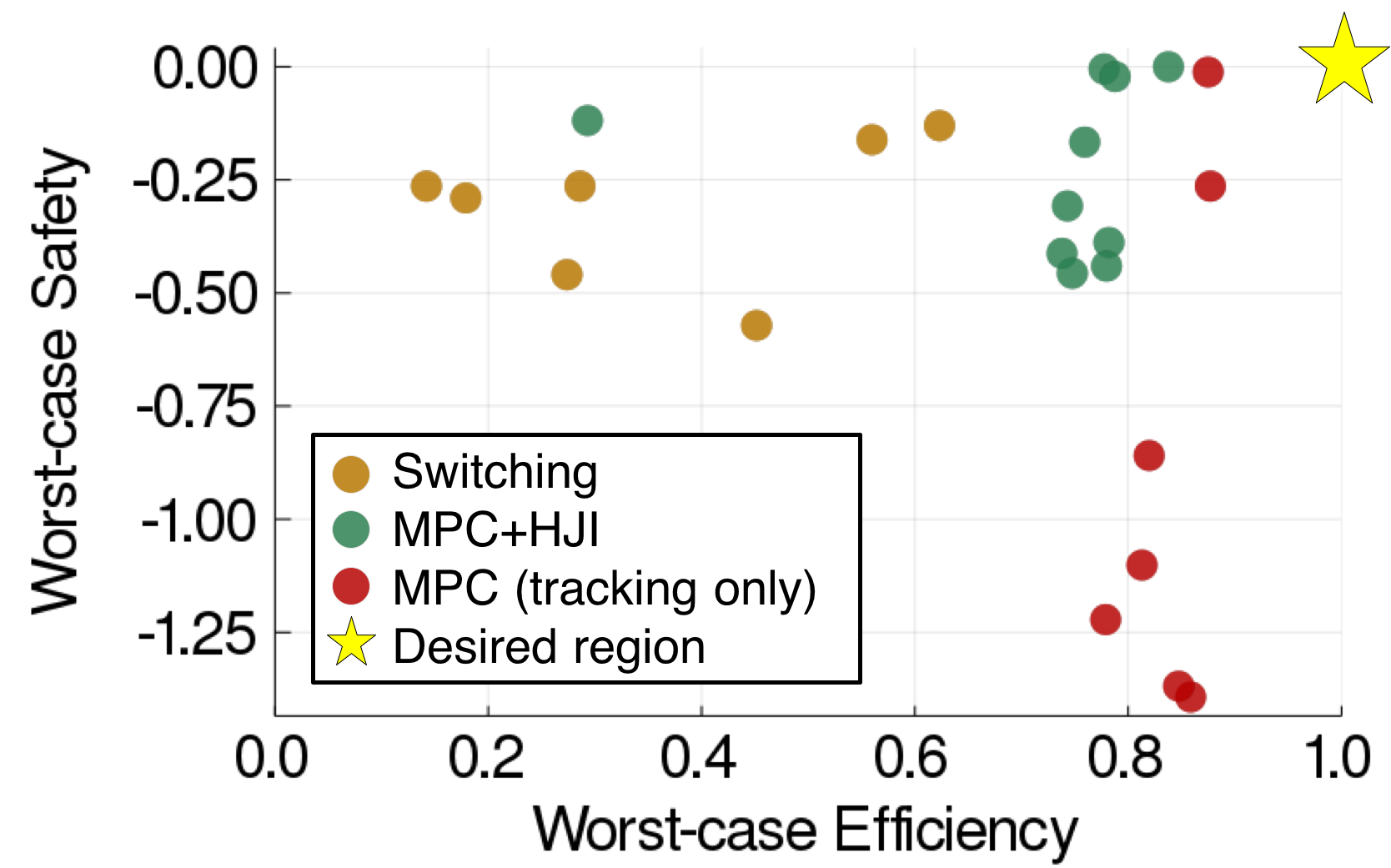}}
\caption{Safety and efficiency trade-offs for the different control strategies.}
\label{fig:tradeoff}
\end{figure}

\subsection{Experiments}\label{sec:experiments}

To evaluate our proposed control stack---a synthesis of a high-level probabilistic interaction planner with the MPC+HJI tracking controller---we perform full-scale human-in-the-loop traffic-weaving trials with X1 taking on the role of the robot car. To ramp up towards testing with two full-scale vehicles in the near future, we investigate two types of human car: (i) a virtual human-driven car and (ii) a 1/10-scale LiDAR-visible human-driven RC car. We scale the highway traffic-weaving scenario (mean speed $\sim$28m/s) in \cite{SchmerlingLeungEtAl2018} down to a mean speed of $\sim$8m/s by shortening the track (reducing longitudinal velocity by a constant) and scaling time by a factor of 4/3 (with the effect of scaling speeds by 3/4 and accelerations by 9/16). The value of the parameters in the MPC tracking problem are listed in Appendix \ref{app:trajectory parameters}.

We investigate and evaluate the effectiveness of our control stack by allowing the human car to act carelessly (i.e., swerving blindly towards the robot car) during the experiments. In Sections~\ref{sec:experiments_comparison} and \ref{sec:experiments_safety} we compare our proposed controller (MPC+HJI) against a tracking-only MPC controller (MPC) and a controller that switches to the HJI optimal avoidance controller when near safety violation (switching). In Section~\ref{sec:static wall result} we compare the two static obstacle avoidance approaches developed in Section~\ref{sec:wall_def} to a scenario with a virtual wall constraining the robot's trajectory. We investigate applying an extension of the same state-constraint-based static obstacle avoidance strategy to the problem of collision with a dynamic obstacle (the human) in Section~\ref{sec:experiments_baseline} and discuss its shortcomings. Finally, in Section~\ref{sec:experiments_rccar} we present experiments with the human car physically embodied by an RC car.

\subsubsection{Virtual Human-Driven Vehicle}\label{sec:experiments_comparison}

To ensure a completely safe experimental environment, our first tier of experimentation uses a joystick-controlled virtual vehicle for the human car and allows the robot control stack to have perfect observation of the human car state.
Experimental trials of the probabilistic planning framework using (i) our proposed approach (MPC+HJI) and (ii) switching to the optimal HJI controller (switching) are shown in Figure~\ref{fig:controller comparison}, along with a simulated comparison between the two safety controllers and the tracking-only controller (MPC). For comparative purposes, the controllers were simulated with the displayed nominal trajectory held fixed, but in reality, the nominal trajectory in these experiments was updated at $\sim 3Hz$.
As expected, we see that when safety violation occurs, the MPC+HJI controller represents a middle ground between the tracking-only MPC which does not react to the human car's intrusion, and the switching controller which arguably overreacts with a large excursion outside the lane boundaries.
Evidently, our proposed controller tries to be minimally interventional---the robot car swerves/brakes but only to an extent that is necessary.

Looking at the value function, we see that, as expected, the MPC+HJI controller aims to keep the value positive, but does not necessarily strive to increase it, while the switching controller aims to increase the value as much as possible. The MPC (tracking only) controller fails to increase the value at all when safety violation occurs. All controllers however, experience a period where the value is negative, even the two HJI-based controllers which theoretically guarantee safety. We believe this is due to model mismatch; we will discuss this point in more depth in Section \ref{sec:insights}.

Additionally, in some trials when the robot car was traveling faster, the activation of the HJI constraint resulted in the robot car performing a large but smooth swerve that traversed completely outside the lane boundaries. We address this limitation by adding a wall constraint into the MPC formulation discussed in Section \ref{sec:static wall result}.

\subsubsection{Safety and Efficiency Trade-off}\label{sec:experiments_safety}

Our MPC+HJI controller considers the set of safety-preserving controls while optimizing its tracking performance when it is near safety violation. In contrast, the HJI switching controller uses the optimal avoidance control policy and as a result neglects to track the desired trajectory that was selected by the planner for interaction performance. As such, there is a trade off between safety, defined with respect to the value function, and efficiency.

For use as a comparison metric, we define the notion of \emph{total safety} of an interaction of time length $T$ as
\[S_\mathrm{total} := \int_0^T\; \mathbf{1}[V(\Xrel(t)) \leq 0] V(\Xrel(t)) \;dt\]
which is the integral of the value function when it is negative. Total safety considers not only the magnitude of the safety violation, but also the duration of the violation. We can also define the \emph{worst-case safety} as
\[ S_\mathrm{worst}:=\min_{t\in[0,T]} V(\Xrel(t))\]
which does not consider the duration of safety violation, but rather the worst-case safety violation with respect to the value function over the interaction.

Efficiency of the interaction is more difficult to quantify. In this analysis, we define efficiency with respect to the $g$-forces experienced by the vehicle as this is a proxy for control effort and passenger comfort. Alternative metrics include time taken to complete the interaction, and friction available in the tires. We define \emph{average efficiency} as
\[ E_\mathrm{avg}:=1 - \frac{1}{T}\int_0^T  \frac{1}{g}\sqrt{a_x(t)^2 + a_y(t)^2} \;dt \]
where $a_x(t)$ and $a_y(t)$ are the $x$ and $y$ acceleration of the robot car; larger $E_\mathrm{avg}$values correspond to better efficiency. $g$-forces should not exceed one as this is beyond the physical limits of a vehicle. We also define the \emph{worst-case efficiency} as
\[ E_\mathrm{worst}:=\max_{t\in[0,T]} \left(1 - \frac{1}{g}\sqrt{a_x(t)^2 + a_y(t)^2}\right)\]
which considers the largest $g$-force experienced during the interaction.

Given these quantities, we can compare the safety and efficiency trade-offs between the MPC, MPC+HJI, and switching controllers. Multiple experimental trials using X1 were carried out using each controller, and Figure \ref{fig:tradeoff average} compares the average/total metrics while Figure \ref{fig:tradeoff worst-case} compares the worst-case metrics. We see that in both cases, the MPC+HJI controller provides a good balance between safety and efficiency; the safety almost equaling that of the optimal switching controller, and efficiency almost equaling that of the MPC controller.
The switching controller provides the highest level of safety (with respect to the value function) but experiences lower efficiency since it often results in heavy braking and sharp swerving. The MPC controller provides lower safety scores but with larger variations as the resulting safety score is scenario dependent rather than controller dependent.

\begin{figure}[th]
    \centering
    \subfloat[Comparison of steering angle. The plot begins when the robot car is planning autonomously.]{
    \label{fig:wall comparison delta}
    \includegraphics[width=0.45\textwidth]{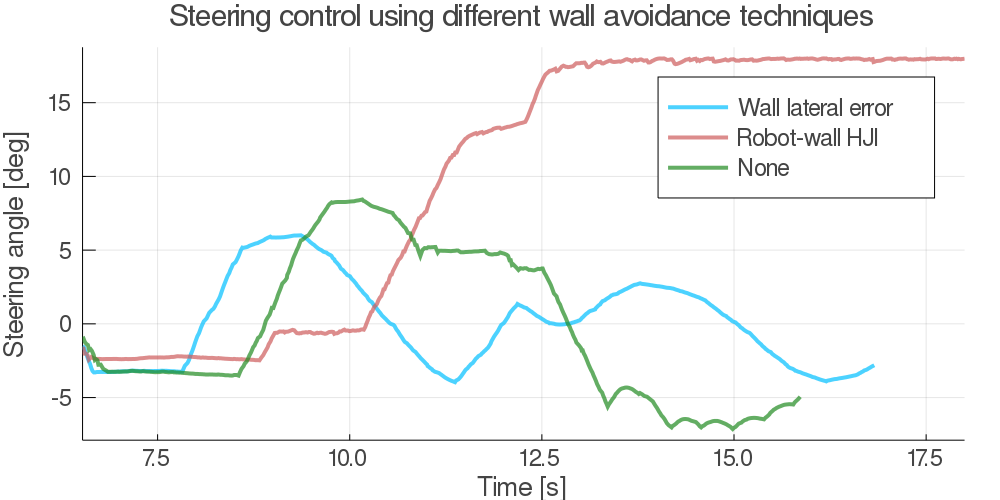}
    }\\
    \subfloat[Comparison of longitudinal force. The plot begins when the robot car is planning autonomously.]{
    \label{fig:wall comparison Fx}
    \includegraphics[width=0.45\textwidth]{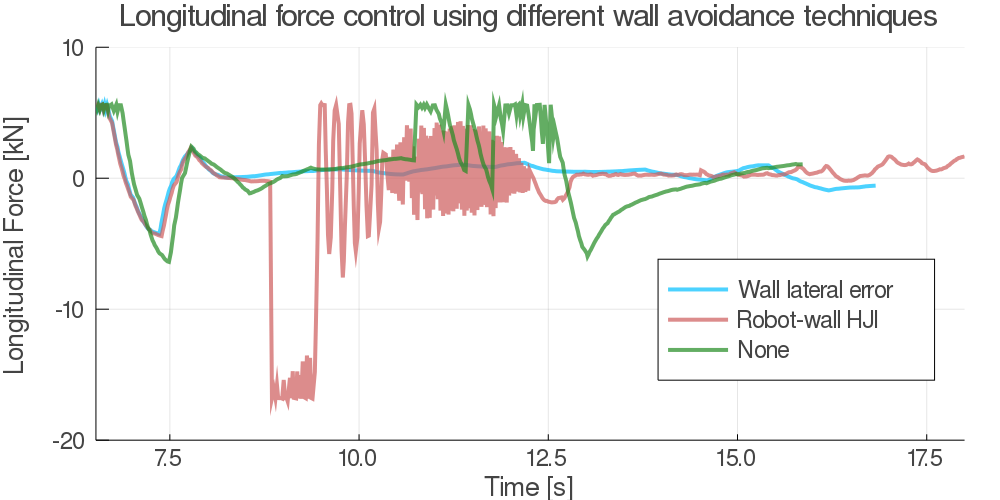}
    }
    \caption{Comparison of control sequences from using MPC lateral error state constraints, robot-wall HJI control} constraint, and no constraints to avoid a static road boundary.
    \label{fig:wall comparison}
\end{figure}

\begin{figure*}[tb]
    \centering
    \includegraphics[width=\textwidth]{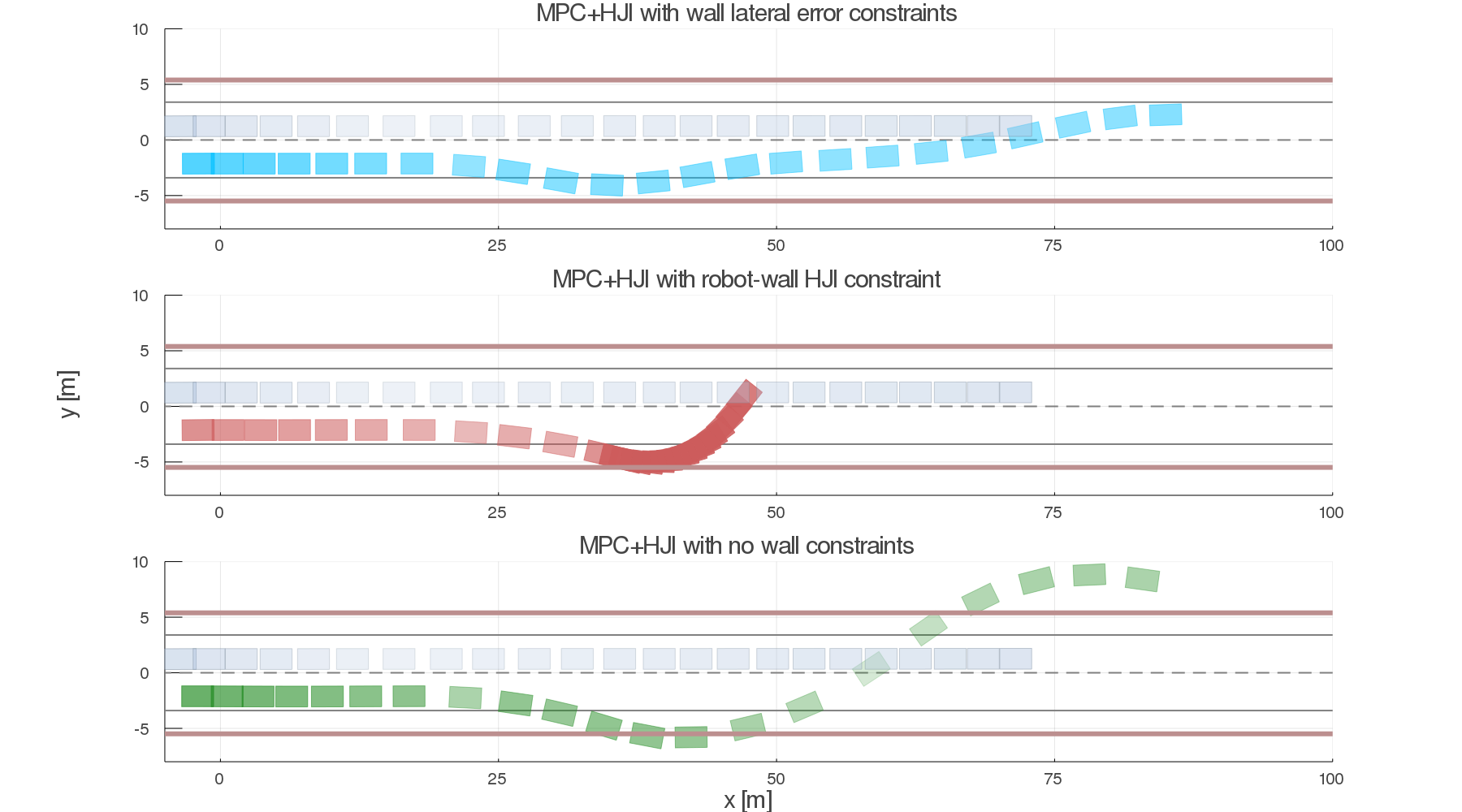}
    \caption{Trajectory comparison between using lateral error constraints (top), a robot-wall HJI control constraint (middle), and using no wall constraints (bottom). The human car is in the left (top) lane and is constant across all three trials, and the robot car is in the right (bottom) lane. The transparency of the car corresponds to the speed of the car (higher transparency corresponds to higher speed).}
    \label{fig:wall comparison trajectory}
\end{figure*}

\subsubsection{Static wall}
\label{sec:static wall result}

We investigate two methods, (i) persistently adding lateral error state constraints into the MPC problem and (ii) adding an additional HJI-based control constraint for a robot-wall relative system into the MPC problem when $V_{\WALL}(\xrob) \leq \epsilon$, for avoiding a static (virtual) road boundary wall when the robot car is swerving off the road to avoid a collision with the human car.

Figure \ref{fig:wall comparison} compares the control sequences and Figure \ref{fig:wall comparison trajectory} compares the trajectory of both approaches. Here, the human car is commanded to stay in the lane and is initialized inside the BRS so that when the robot starts planning autonomously, the robot car will react immediately and swerve out of the way---we investigate the nature of the swerving. We do note that the human car is able to make the robot car swerve even though it is staying in its lane. This is due to the wide, pear-shape BRS when the two cars are parallel and at similar speeds (see Figure \ref{fig:BRS illustration} left). We discuss this in more detail in Section~\ref{sec:insights}.

Around 20 meters from the start of the roadway in Figure \ref{fig:wall comparison trajectory}, the robot car begins to plan autonomously. Before then, the cars are following a straight line in order to speed up from $\approx 1m/s$ to the desired interaction speed. To provide the cleanest comparison, all results in this subsection are derived from simulation.
Starting in the right lane, we see in all cases that when safety is violated the robot car swerves to the right ($\delta < 0$).
When using the lateral error constraints (Figure \ref{fig:wall comparison trajectory}, top), the robot car essentially has a ``look-ahead'' capability because it is able to optimize its steering commands over the MPC tracking horizon, essentially distributing the responsibility of avoiding the wall across the entire tracking MPC horizon. As a result, the robot car is able to successfully and smoothly steer back onto the road and avoid the wall.

When using the robot-wall HJI control constraint (Figure \ref{fig:wall comparison trajectory}, center), the robot car does not preemptively swerve back onto the road and instead only reacts to the wall when $V_{\WALL}(\xrob) \leq \epsilon$ as designed. The robot car performs a hard brake to the point of almost stopping\footnote{Since the dynamic bicycle model is ill-defined for low speeds (thus explaining the oscillations in $\Fx$), the experiment (when using the robot-wall HJI control constraint) is essentially over around $t=9$. In general, the MPC problem can switch to the kinematic model at low speeds \citep{PattersonThorntonEtAl2018} which is well defined in that speed region.} and then begins to eventually command a maximum steering angle (18 deg). The sharp and abrupt behavior stems from the HJI formulation assuming the robot can and will take extreme actions, including cases when using the safety-preserving control set.
As a result, the responsibility for evasive action is triggered at the very latest possible instance and compressed into a control constraint over a single (in practice 3) MPC timestep. This is in contrast to the other approach of always having MPC lateral error state constraints over the entire MPC horizon.
The MPC+HJI controller is effective in avoiding dynamic obstacles, but for static obstacles, HJI is not suitable because it is unnecessary to reason about the dynamics of something that is static.
With no wall constraints (Figure \ref{fig:wall comparison trajectory}, bottom), the robot car swerves completely off the road, and in order to quickly get to the left lane before the end of the road (inscribed as an objective in the high-level planner), it overshoots to the other side of the road and also drives beyond the road boundary on the other side.

\begin{figure*}[ht]
\centering
\subfloat[Simulation trial of using MPC+HJI for human car collision avoidance (including lateral error constraints for wall collision avoidance).]{
    \label{fig:dynamic obstacles mpc hji traj}
    \includegraphics[width=0.48\textwidth]{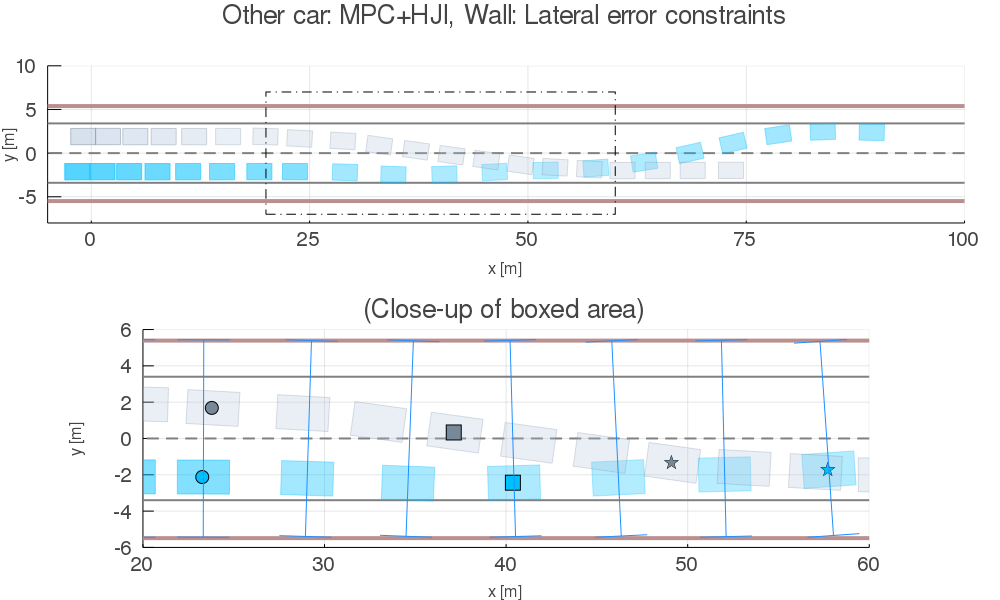} }
\subfloat[Simulation trial of using lateral error constraints for both human car and wall collision avoidance.]{
    \label{fig:dynamic obstacles lateral error traj}
    \includegraphics[width=0.48\textwidth]{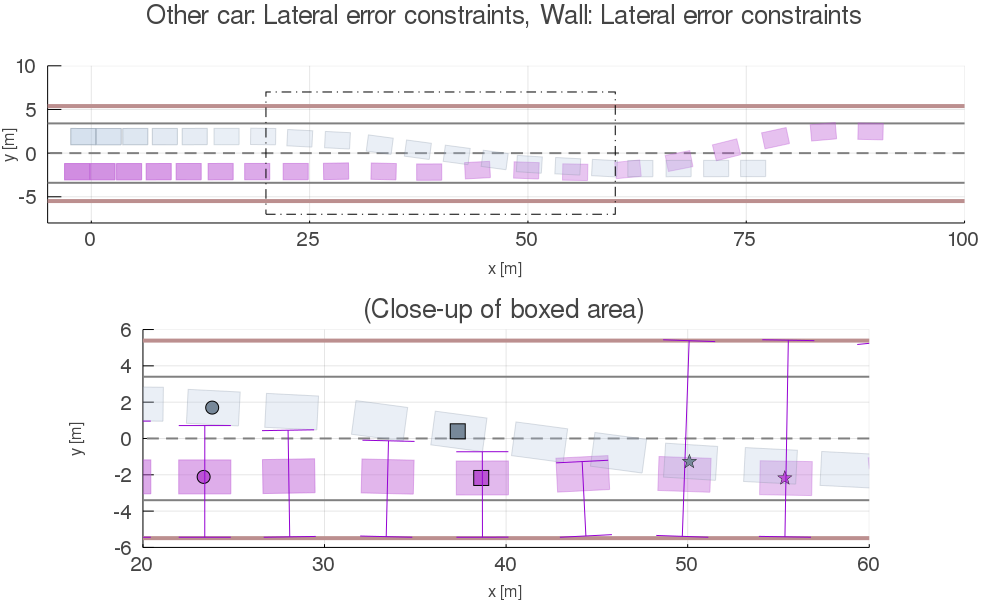}}
\caption{Comparison between the MPC+HJI controller (left) and a baseline approach not based on any reachability analysis (right) for the task of dynamic obstacle avoidance. The trajectory of the human car starting in the left (top) lane is visualized in gray; the trajectory of the robot car starting in the right (bottom) lane is visualized in color.
In each close-up view, error bars accompanying each snapshot of the robot illustrate the first ($k=1$) lateral state constraint of the MPC problem at that step.
To help correspond the positions of the cars over the trials, each pair of matching-shape markers represents a common time instant.
}
\label{fig:dynamic obstacle traj}
\end{figure*}

\subsubsection{Comparison to baseline dynamic obstacle avoidance}\label{sec:experiments_baseline}

Motivated by the success of applying state constraints in the previous section, we compare our proposed framework (MPC+HJI for safe human-robot interaction) with a baseline approach of avoiding collision with the human car using lateral error constraints. In order to define safety state constraints over the duration of the robot trajectory, we must however make some assumption on the human's future trajectory. For this baseline, at each MPC iteration, we assume that the human car will continue moving at its current heading and with its current velocity.\footnote{Alternatively, one could use a sample or a \emph{maximum a posteriori} estimate from the interaction planner's prediction model.} The lateral error constraint at each MPC time step is computed between the robot's desired trajectory and the human's projected trajectory---this is analogous to applying the approach taken in \cite{BrownFunkeEtAl2017} but with (deterministic) dynamic obstacles. A simulation of our proposed approach and the baseline approach is shown in Figure \ref{fig:dynamic obstacle traj}. Notably, in this simulation, the human car defies the baseline's linear-extrapolation-based prediction model and curves into the other lane towards the robot car.

We see in Figure~\ref{fig:dynamic obstacles mpc hji traj} that when using the MPC+HJI controller (with lateral error state constraints for wall avoidance), we are able to successfully avoid a collision with the human car and the wall, and maintain relative states outside the BRS. The baseline approach demonstrated in Figure~\ref{fig:dynamic obstacles lateral error traj}, on the other hand, collides with the human because its consideration of only one possible future (i.e., human trajectory) at each MPC step leads the robot into a region of inevitable collision (defined against all, including worst-case, human controls). In particular we see that as the human trajectory increases its curvature towards the middle of the lane change, the lateral error bounds (based on an assumption of constant velocity) are not stringent enough to maintain robot safety. The utility of the MPC+HJI approach is that it distills safety considerations with respect to all possible realizations of the human trajectory into a single control constraint. Moreover, this constraint is not overly conservative (to improve the state-constraint-based approach one might imagine defining lateral error constraints to avoid the entire forward reachable set of the human), because it incorporates the concept of closed-loop feedback into its BRS computation.

\begin{figure*}[t]
    \centering
    \includegraphics[width=\textwidth]{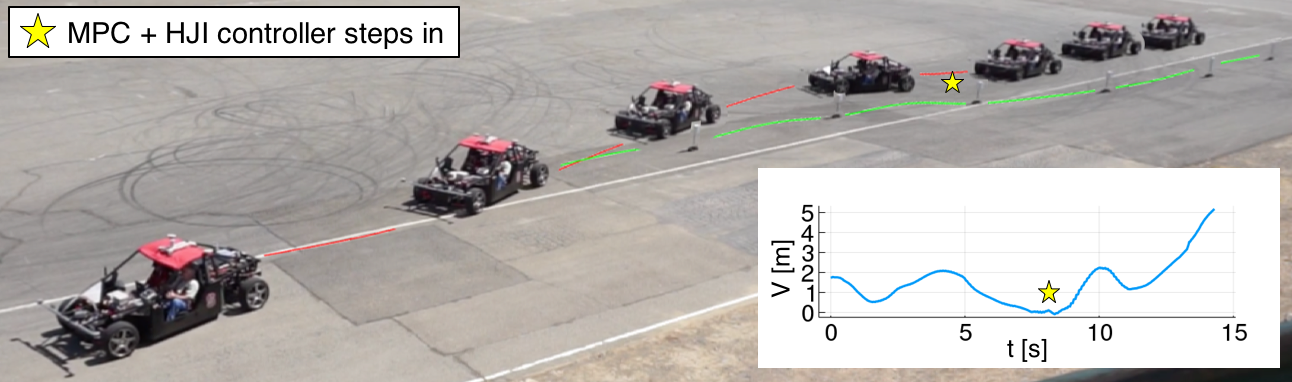}
    \caption{Time-lapse of pairwise vehicle interaction: X1 with 1/10-scale human-driven RC car. The RC car (green trajectory) nudges into X1 (red trajectory), which swerves gently to avoid. Inset: The value function over time of the X1-human driven RC car experiment}
    \label{fig:rccar time lapse}
\end{figure*}

\subsubsection{1/10-Scale Human-Driven Vehicle}\label{sec:experiments_rccar}
To begin investigating the effects of perception uncertainty on our safety assurance framework, we use three LiDARs onboard X1 to track a human-driven RC car, and implement a Kalman filter for human car state estimation (position, velocity, and acceleration). Even with imperfect observations, we show some successful preliminary results (an example is shown in Figure~\ref{fig:rccar time lapse}) at mean speeds of ~4m/s, close to the limits of the RC car + LiDAR-visible mast in crosswinds at the test track. We observe similar behavior as in the virtual human car experiments, including the fact that the value function dips briefly below zero before the MPC+HJI controller is able to arrest its fall; we discuss this behavior in the next section.

\section{Discussion}\label{sec:insights}

Beyond the qualitative and quantitative confirmation of our design goals, our experimental results reveal three main insights.

\vspace{.1cm}
\noindent \emph{Takeaway 1: The reachability cache is underly-conservative with respect to robot car dynamics and overly-conservative with respect to human car dynamics.}
\vspace{.1cm}

\noindent In all cases---hardware experiments as well as simulation results---the HJI value function $V$ dips below zero, indicating that neither the HJI+MPC nor even the optimal avoidance switching controller are capable of guaranteeing safety in the strictest sense. The root of this apparent paradox is in the computation of the reachability cache used by both controllers as the basis of their safety assurance. Though the 7-state relative dynamics model (Equation \eqref{eqn:relative dynamics}) subsumes a single-track vehicle model that has proven successful in predicting the evolution of highly dynamic vehicle maneuvers \citep{FunkeBrownEtAl2017,BrownFunkeEtAl2017}, the way it is employed in computing the value function $V$ omits relevant components of the dynamics. In particular, when computing the optimal avoidance control (Equation \eqref{eqn: optimal control}) as part of solving the HJI PDE, we assume total freedom over the choice of robot steering angle $\delta$ and longitudinal force command $F_x$, %
up to maximum control limits. This does not account for, e.g., limits on the steering slew rate (traversing $[\delta_{\min}, \delta_{\max}]$ takes approximately 2 seconds), and thus the value function is computed under the assumption that the robot can brake/swerve far faster than it actually can.

We note that simply tuning the safety buffer $\epsilon$ is insufficient to account for these unmodeled dynamics. In Figure~\ref{fig:HJI+MPC comparison} we see that $V$ may drop from approximately 0.5 (the value when the two cars traveling at 8m/s start side-by-side in lanes) to -0.3 in the span of a few tenths of a second. Selecting $\epsilon > 0.5$ might give enough time for the steering to catch up, but such a selection would prevent the robot car from accomplishing the traffic weaving task even under nominal conditions, i.e., when the human car is equally concerned about collision avoidance. Even for $\epsilon = 0.05$, we see that the human car can cause the robot to swerve by just staying in its lane but with a small offset towards the robot's lane (see Figure \ref{fig:wall comparison trajectory}).
This is because the safety controller would push the robot car outside of its lane from the outset to maintain the buffer. This behavior follows from wide level sets associated with the transient control authority asymmetry (recall that in the HJI relative dynamics the human car may adjust its trajectory curvature discontinuously), assumed as a conservative safety measure as well as a way to keep the relative state dimension manageable.

The simplest remedy for both of these issues is to increase the fidelity of the relative dynamics model by incorporating additional integrator states $\dot\delta$ for the robot and $\dot \omega$ for the human. Naively increasing the state dimension to 8 or 9, however, might not be computationally feasible (even offline) without devising more efficient HJI solution techniques or choosing an extremely coarse discretization over the additional states. By literature standards we already use a relatively coarse discretization grid for solving the HJI PDE; associated numerical inaccuracies may be another source of the observed safety mismatch and alternate grid choices are a possible subject of future investigation. We believe that simulation, accounting for slew rates, could be a good tool to prototype such efforts, noting that as it stands we have relatively good agreement between simulation and the experimental platform in our testing.

\vspace{.1cm}
\noindent\emph{Takeaway 2: Interpretability of the value function $V$ should be a key consideration in future work.}
\vspace{.1cm}

\noindent In this work the terminal value function $V(0, \Xrel)$ is specified as the separation/penetration distance between the bounding boxes of the two vehicles, a purely geometric quantity dependent only on $\xrel$, $\yrel$, and $\psirel$. Recalling that V ($:= V_\infty$) represents the worst-case eventual outcome of a differential game assuming optimal actions from both robot and human, we may interpret the above results through the lens of worst-case outcomes, i.e., a value of -0.3 may be thought of as 30cm of collision penetration assuming optimal collision seeking/avoidance from human/robot.
When extending this work to cases with environmental obstacles (e.g., concrete highway boundaries that preclude large deviations from the lane), or multi-agent settings where the robot must account for the uncertainty in multiple other parties' actions, for many common scenarios it may be the case that guaranteeing absolute safety is impossible. Instead of avoiding a BRS of states that might lead to collision, we should instead treat the value function inside the BRS as a cost. In particular we should specify more contextually relevant values of $V(0, \Xrel)$ for states in collision, e.g., negative kinetic energy or another notion of collision severity as a function of the velocity states $\Uxr$, $\Uyr$, $\vh$, and $\rr$ in addition to the relative pose. This would lead to a controller that prefers, in the worst case, collisions at lower speed, or perhaps ``glancing blows'' where the velocities of the two cars are similar in magnitude and direction.

\vspace{.1cm}
\noindent\emph{Takeaway 3: Static obstacles should be accounted for using MPC tracking state constraints and not HJI control constraints.}
\vspace{.1cm}

It is natural to use HJI when reasoning about potential dangers from a dynamic, unpredictable agent, but for static obstacles there is no uncertainty in how the obstacle may act. That is, when the other agent is deterministic, there is no notion of optimizing robot safety policies over all worst-case actions by the other agent. Indeed, in this case it is possible to confidently extend the safety constraint along the entire MPC horizon, as demonstrated in Section~\ref{sec:static wall result}. This is as opposed to deriving a safety constraint to be applied at the first step (on the present control action; the HJI+MPC approach), as must be done with a nondeterministic agent with unknown future trajectory. Spreading the collision avoidance constraint over the problem horizon allows for the MPC optimization to more smoothly accomplish its objectives (i.e., control smoothness and tracking performance) while maintaining safety. We note, however, that naive extensions of this approach to nondeterministic agents (recall Section \ref{sec:experiments_baseline}) do not similarly provide strong safety assurances.

\section{Conclusions}
We have investigated a control scheme for providing real-time safety assurance to underpin the guidance of a probabilistic planner for human-robot vehicle-vehicle interactions. By essentially projecting the planner's desired trajectory into the set of safety-preserving controls whenever safety is threatened, we preserve more of the planner's intent than would be achieved by adopting the optimal control with respect to separation distance. Our experiments show that with our proposed minimally interventional safety controller, we accomplish the high level objective (traffic weaving) despite the human car swerving directly onto the path of the robot car, and accomplish this relatively smoothly compared to using a switching controller that results in the robot car swerving more violently off the road.
Further, we investigate the addition of a road boundary wall to our formulation to prevent the robot car from swerving completely out of the lane which could be dangerous in realistic road settings, and compare against a baseline approach using state constraints to avoid dynamic obstacles to show that our framework is better designed to keep the robot car safe even in the face of worst-case human car behaviors.

We note that this work represents only a promising first step towards the integration of reachability-based safety guarantees into a probabilistic planning framework.
Future work includes investigating this framework for
cases where the human and robot have very different dynamics, such as a pedestrian or cyclist interacting with a car, or a human interacting with a robotic manipulator, and for interactions involving multiple (more than two) agents. We may also consider adapting our approach to ensure high planning performance while guaranteeing satisfaction of constraints other than safety, e.g., task requirements specified by temporal logic constraints as considered in \cite{ChenTamEtAl2018}.
In the context of human-robot vehicle interactions, we have already discussed the concrete modifications to this controller we believe are necessary to improve the practical impact of our theoretical guarantees; further study should also consider better fitting of the planning objective at the controller level. That is, instead of performing a naive projection, i.e., the one that minimizes trajectory tracking error, it is likely that a more nuanced selection informed by the planner's prediction model would represent a better ``backup choice'' in the case that safety is threatened.
We recognize that ultimately, guaranteeing absolute safety on a crowded roadway may not be realistic, but we believe that in such situations value functions derived from reachability may provide a useful metric for near-instantly evaluating the future implications of a present action choice.

\section*{Acknowledgements}
This work was supported by the Office of Naval Research (Grant N00014-17-1-2433), by Qualcomm, and by the Toyota Research Institute (``TRI''). This article solely reflects the opinions and conclusions of its authors and not ONR, Qualcomm, TRI, or any other Toyota entity. The authors would like to thank the X1 team, in particular Matt Brown, Larry Cathey, and Amine Elhafsi, Matteo Zallio, and the Thunderhill Raceway Park for accommodating testing.

\bibliographystyle{SageH}
\bibliography{main,ASL_papers}

\newpage
\appendix

\section{Vehicle Parameters} \label{app:vehicle parameters}
\begin{table}[h]
    \caption{X1 parameters relevant to defining the equations of motion in Equation \eqref{eqn:bicycle model}.}
    \label{tab:vehicle parameters}
    \centering
    \begin{tabular}{|c|c|c|}
    \hline
    \hline
    \textbf{Variable} & \textbf{Description} & \textbf{Value} \\
    \hline
    \hline
        $g$ & Standard Earth gravity & 9.80665$\mathrm{ms}^{-2}$\\
        $m$ & Total mass & 1964kg\\
        $I_{zz}$ & Yaw moment of inertia & 2900 kgm$^2$\\
        $h$& Vertical height of CG  & 0.47m\\
        $d_{_\mathrm{f}}$ & Distance from CG to front axle  & 1.4978m\\
        $d_{_\mathrm{r}}$& Distance from CG to rear axle & 1.3722m\\
        $C_{\alpha_f}$& Front cornering stiffness & 150kN/rad\\
        $C_{\alpha_r}$& Rear cornering stiffness & 220kN/rad\\
        $C_{d_0}$& Drag coefficient constant & 241N\\
        $C_{d_1}$& Drag coefficient linear & 25.1Nm$^{-1}$s\\
        $C_{d_2}$& Drag coefficient  quadratic & 0.0Nm$^{-2}$s$^2$\\
        $\Fxf$ & Front wheel drive fraction ($\Fx \geq 0$) & 0\\
        $\Fxr$ & Rear wheel drive fraction ($\Fx \geq 0$) & 1\\
        $\Fxf$ & Front wheel brake fraction ($\Fx < 0$) & 0.6\\
        $\Fxr$ & Rear wheel brake fraction ($\Fx < 0$) & 0.4\\
    \hline
    \end{tabular}

\end{table}

\section{Trajectory Parameters}\label{app:trajectory parameters}

\begin{table}[h]
    \caption{Parameter values relevant for the MPC tracking optimization in Equation \eqref{eqn:QP}.}
    \label{tab:MPC parameters}
    \centering
    \begin{tabular}{|c|p{0.3\textwidth}|p{0.1\textwidth}|}
    \hline

    \hline
    \textbf{Variable} & \textbf{Description} & \textbf{Value} \\
    \hline
    \hline
        $Q_{\Delta s}$ & Quadratic cost on longitudinal error & 1.0m$^{-2}$s$^{-1}$\\
        $Q_{\Delta \psi}$ & Quadratic cost on heading error & 1.0s$^{-1}$\\
        $Q_{e}$ & Quadratic cost on lateral error  & 1.0m$^{-2}$s$^{-1}$\\
        $R_{\Delta  \delta} $ & Quadratic cost on change in steering angle & 0.1s\\
        $R_{\Delta \Fx}$ & Quadratic cost on change in longitudinal tire force & 0.5N$^{-2}$s\\
        $W_{\beta}$ & Linear cost on sideslip stability slack variable & $\frac{900}{\pi}$s$^{-1}$\\
        $W_r$ & Linear cost on yaw rate stability slack variable & 50.0\\
        $W_{\HJI}$ & Linear cost on HJI slack variable & 500.0m$^{-1}$\\
        $W_e$ & Linear cost on lateral bound slack variable & 500.0m$^{-1}$s$^{-1}$\\
        $\delta_{\min}$ & Minimum steering angle & -18$\times \frac{\pi}{180}$\\
        $\delta_{\max}$ & Maximum steering angle & 18$\times \frac{\pi}{180}$\\
        $\dot{\delta}_{\min}$ & Minimum steering rate& -0.344 $s^{-1}$\\
        $\dot{\delta}_{\max}$ & Maximum steering rate & 0.344 $s^{-1}$\\
        $F_{x,\min}$ & Minimum longitudinal tire force & -16794N\\
        $F_{x,\max}$ & Maximum longitudinal tire force & $\min(5600,$ $\frac{75000}{U_x})$N\\
        $U_{x,\min}$ & Minimum longitudinal velocity & 1.0ms$^{-1}$\\
        $U_{x,\max}$ & Maximum longitudinal velocity & 15ms$^{-1}$\\
        $N_\mathrm{long}$ & Number of long MPC time steps & 10\\
        $N_\mathrm{short}$ & Number of short MPC time steps & 5\\
        $N_\HJI$ & Number of time steps with HJI constraint applied& 3\\
        $\Delta t_\mathrm{long}$ & Length of long MPC time step & 0.2s\\
        $\Delta t_\mathrm{short}$ & Length of short MPC time step & 0.01s\\
        $\epsilon$ & Size of HJI value function buffer & 0.05m\\
    \hline
    \end{tabular}

\end{table}

\end{document}